%% file: icml2023.tex
\documentclass[nohyperref]{article}

\usepackage[table, dvipsnames]{xcolor}
\usepackage{microtype}
\usepackage{graphicx}
\usepackage{subfigure}
\usepackage{booktabs} 

\usepackage{hyperref}

\usepackage[accepted]{icml2023}

\usepackage{amsmath}
\usepackage{amssymb}
\usepackage{mathtools}
\usepackage{amsthm}

\usepackage[capitalize,noabbrev]{cleveref}

\theoremstyle{plain}
\newtheorem{theorem}{Theorem}[section]
\newtheorem{proposition}[theorem]{Proposition}

\theoremstyle{definition}

\theoremstyle{remark}

\input{math_commands}

\usepackage{xspace}
\usepackage{lipsum}
\usepackage{threeparttable}
\usepackage{tablefootnote}

\allowdisplaybreaks

\newcommand{\cx}[1]{\mathbf{x}_{#1}}
\newcommand{\ptheta}{p_\theta}
\newcommand{\qphi}{q_\phi}
\newcommand{\bb}[1]{{\mathbb{#1}}}
\newcommand{\norm}[1]{{\lVert {#1} \rVert}}

\newcommand{\method}{\textsc{GeoLDM}\xspace}

\usepackage[textsize=tiny]{todonotes}

\icmltitlerunning{Geometric Latent Diffusion Models for 3D Molecule Generation}

\begin{document}

\twocolumn[
\icmltitle{Geometric Latent Diffusion Models for 3D Molecule Generation}



\icmlsetsymbol{equal}{*}

\begin{icmlauthorlist}
\icmlauthor{Minkai Xu}{cs}
\icmlauthor{Alexander S. Powers}{cs,chem}
\icmlauthor{Ron O. Dror}{equal,cs}
\icmlauthor{Stefano Ermon}{equal,cs}
\icmlauthor{Jure Leskovec}{equal,cs}
\end{icmlauthorlist}

\icmlaffiliation{cs}{Department of Computer Science, Stanford University}
\icmlaffiliation{chem}{Department of Chemistry, Stanford University}

\icmlcorrespondingauthor{Minkai Xu}{minkai@cs.stanford.edu}

\icmlkeywords{Machine Learning, ICML}

\vskip 0.3in
]



\printAffiliationsAndNotice{*Equal senior authorship} 

\begin{abstract}
    Generative models, especially diffusion models (DMs), have achieved promising results for generating feature-rich geometries and advancing foundational science problems such as molecule design. Inspired by the recent huge success of Stable (latent) Diffusion models, we propose a novel and principled method for 3D molecule generation named Geometric Latent Diffusion Models (\method). \method is the first latent DM model for the molecular geometry domain, composed of autoencoders encoding structures into continuous latent codes and DMs operating in the latent space. Our key innovation is that for modeling the 3D molecular geometries, we capture its critical roto-translational equivariance constraints by building a point-structured latent space with both invariant scalars and equivariant tensors. Extensive experiments demonstrate that \method can consistently achieve better performance on multiple molecule generation benchmarks, with up to 7\% improvement for the valid percentage of large biomolecules. Results also demonstrate \method's higher capacity for controllable generation thanks to the latent modeling. Code is provided at \url{https://github.com/MinkaiXu/GeoLDM}.
\end{abstract}

\section{Introduction}

Generative modeling for feature-rich geometries is an important task for many science fields. 
Typically, geometries can be represented as point clouds where each point is embedded in the Cartesian coordinates and labeled with rich features. 
Such structures are ubiquitous in scientific domains, \textit{e.g.}, we can represent molecules as atomic graphs in 3D~\citep{schutt2017schnet} and proteins as proximity spatial graphs over amino acids~\citep{jing2021gvp}. 
Therefore, developing effective geometric generative models holds great promise for scientific discovery problems such as material and drug design~\citep{pereira2016boosting,graves2020review,townshend2021atomd}. 
Recently, considerable progress has been achieved with machine learning approaches, especially deep generative models. 
For example, \citet{gebauer2019symmetry,luo2021autoregressive} and \citet{satorras2021enflow} proposed data-driven methods to generate 3D molecules (in silico) with autoregressive and flow-based models respectively. 
However, despite great potential, the results are still unsatisfactory with low chemical validity and small molecule size, due to the insufficient capacity of the underlying generative models~\citep{razavi2019generating}.

Most recently, diffusion models (DMs)~\citep{ho2020denoising,song2021scorebased} have emerged with surprising results on image synthesis~\citep{meng2022sdedit} and beyond~\citep{kong2021diffwave,li2022diffusionlm}. DMs define a diffusion process that gradually perturbs the data, and learn neural networks to reverse this corruption by progressive denoising. Then the denoising network can conduct generation by iteratively cleaning data initialized from random noise. 
Several studies have also applied such frameworks to the geometric domain, especially molecular structures~\citep{hoogeboom2022equivariant,wu2022diffusionbased,anand2022protein}. 
However, the existing models typically run DMs directly in the atomic feature space, which typically is composed of diverse physical quantities, \textit{e.g.}, charge, atom types, and coordinates. These features are multi-modal with discrete, integer, and continuous variables,
making unified Gaussian diffusion frameworks sub-optimal~\citep{hoogeboom2022equivariant,wu2022diffusionbased} or requiring sophisticated, decomposed modeling of different variables~\citep{anand2022protein}. 
Besides, the high dimensionality of input features also increases DM modeling difficulty, since the model's training and sampling require function forward and backward computation in the full input dimension.
Therefore, the validity rate of generated molecules is still not satisfying enough, and an ideal approach would be a more flexible and expressive framework for modeling complex structures.

\begin{figure*}[!t]
    \centering
    \includegraphics[width=0.9\linewidth]{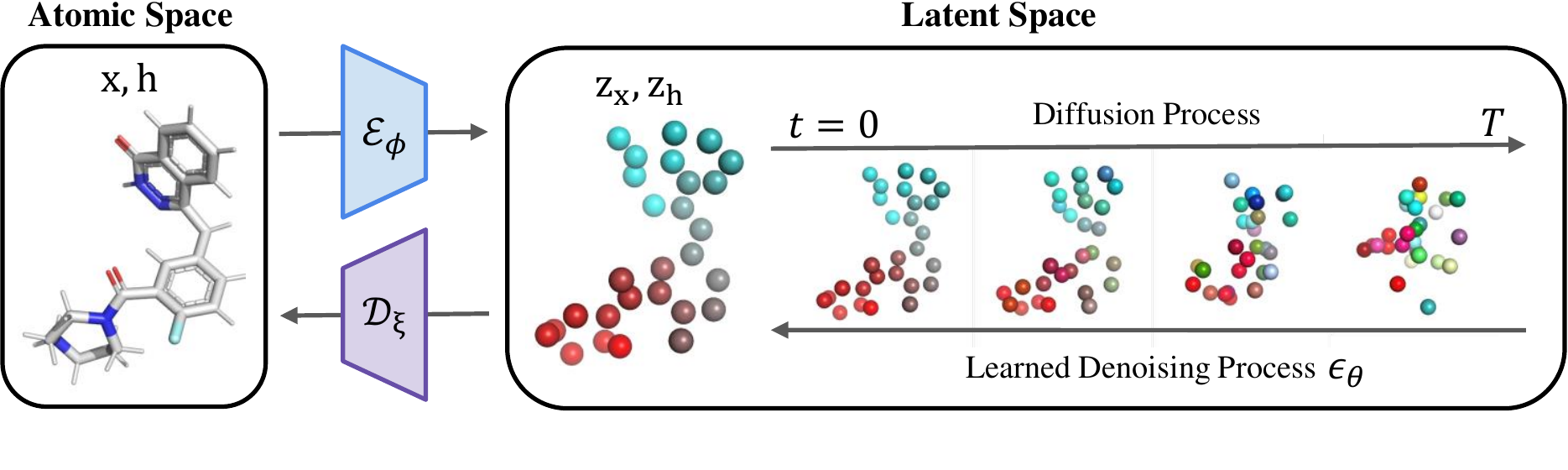} 
    \vspace{-20pt}
    \caption{Illustration of \method. 
    The encoder $\gE_\phi$ encodes molecular features $\rvx, \rvh$ into equivariant latent variables $\rvz_\rx, \rvz_\rh$, and the latent diffusion transitions $q(\rvz_{\rx,t}, \rvz_{\rh,t} | \rvz_{\rx,t-1}, \rvz_{\rh,t-1})$ gradually added noise until the latent codes converge to Gaussians.
    Symmetrically, for generation, an initial latent $\rvz_{\rx,T}, \rvz_{\rh,T}$ is sampled from standard normal distributions and progressively refined by equivariant denoising dynamics $\vepsilon_{\theta}(\rvz_\rx, \rvz_\rh)$. The final latents $\rvz_\rx, \rvz_\rh$ are further decoded back to molecular point clouds with the decoder $\gD_\xi$.}
    \label{fig:framework}
    \vspace{-5pt}
\end{figure*}

In this paper, 
we propose a novel and principled method to overcome the above limitations by utilizing a smoother latent space, named Geometric Latent Diffusion Models (\method). 
\method is set up as (variational) autoencoders (AEs) with DMs operating on the latent space. The encoder maps the raw geometries into a lower-dimensional representational space, and DMs learn to model the smaller and smoother distribution of latent variables. 
For modeling the 3D molecular geometry, our key innovation is constructing sufficient conditions for latent space to satisfy the critical 3D roto-translation equivariance constraints, where simply equipping latent variables with scalar-valued\footnote{In this paper, we will use ``scalar" and ``tensor" to interchangeably refer to type-0 (invariant) and type-1 (equivariant) features, following the common terminologies used in geometric literature.} (\textit{i.e.}, invariant) variables lead to extremely poor generation quality. 
Technically, we realize this constraint by building the latent space as point-structured latents with both invariant and equivariant variables, which in practice is implemented by parameterizing encoding and decoding functions with advanced equivariant networks.
To the best of our knowledge, we are the first work to incorporate equivariant features, \textit{i.e.}, tensors, into the latent space modeling. 

A unique advantage of \method is that unlike previous DM methods operating in the feature domain, we explicitly incorporate a latent space to capture the complex structures. This unified formulation enjoys several strengths. First, by mapping raw features into regularized latent space, the latent DMs learn to model a much smoother distribution. This alleviates the difficulty of directly modeling complex structures' likelihood, and is therefore more expressive. Besides, the latent space enables \method to conduct training and sampling with a lower dimensionality, which can also benefit the generative modeling complexity. Furthermore, the use of latent variables also allows for better control over the generation process, which has shown promising results in text-guided image generation~\citep{rombach2022high}. This enables the user to generate specific types of molecules with desired properties. Finally, our framework is very general and can be extended to various downstream molecular problems where DMs have shown promising results, \textit{i.e.}, target drug design~\citep{lin2022diffbp} and antigen-specific antibody generation~\citep{luo2022antigenspecific}.

We conduct detailed evaluations of \method on multiple benchmarks, including both unconditional and property-conditioned molecule generation. Results demonstrate that \method can consistently achieve superior generation performance on all the metrics, with up to 7\% higher valid rate for large biomolecules. Empirical studies also show significant improvement for controllable generation thanks to latent modeling. All the empirical results demonstrate that \method enjoys a significantly higher capacity to explore the chemical space and generate structurally novel and chemically feasible molecules.

\section{Related Work}

\textbf{Latent Generative Models.}
To improve the generative modeling capacity, a lot of research~\citep{dai2018diagnosing,yu2022vectorquantized} has been conducted to learn more expressive generative models over the latent space. VQ-VAEs~\citep{razavi2019generating} proposed to discretize latent variables and use autoregressive models to learn an expressive prior there. \citet{ma2019flowseq} instead employed flow-based models as the latent prior, with applications on non-autoregressive text generation.  Another line of research is inspired by variational autoencoder's (VAE's) problem that the simple Gaussian priors cannot accurately match the encoding posteriors and therefore generate poor samples, and \citet{dai2018diagnosing,aneja2021contrastive} therefore proposed to use VAEs and energy-based models respectively to learn the latent distribution.
Most recently, several works successfully developed latent DMs with promising results on various applications, ranging from image~\citep{vahdat2021score}, point clouds~\citep{zeng2022lion}, to text~\citep{li2022diffusionlm} generation. Among them, the most impressive success is Stable Diffusion models~\citep{rombach2022high}, which show surprisingly realistic text-guided image generation results.
Despite the considerable progress we have achieved, existing latent generative methods mainly work on latent space only filled with typical \textit{scalars}, without any consideration for equivariance. By contrast, we study the novel and challenging task that latent space also contains equivariant \textit{tensors}.

\textbf{Molecule Generation in 3D.}
Although extensive prior work has focused on generating molecules as 2D graphs \citep{jin2018junction, Liu2019constrained, shi2020graphaf}, interest has recently increased in 3D generation. G-Schnet and G-SphereNet~\citep{gebauer2019symmetry,luo2021autoregressive} employed autoregressive approaches to build molecules by sequential attachment of atoms or molecular fragments. Similar frameworks have also been applied to structure-based drug design~\citep{li2021structure,Pocket2Mol,Powers2022}. However, this autoregressive approach requires careful formulation of a complex action space and action ordering. Other studies utilized atomic density grids, by which the entire molecule can be generated in ``one step" by outputting a density over the voxelized 3D space~\citep{masuda2020generating}. However, these density grids lack the desirable equivariance property and require a separate fitting algorithm. In the past year, DMs have attracted attention for molecule generation in 3D~\citep{hoogeboom2022equivariant, wu2022diffusionbased}, with successful application in downstream tasks like target drug generation~\citep{lin2022diffbp}, antibody design~\citep{luo2022antigenspecific}, and protein design~\citep{anand2022protein,trippe2022diffusion}. However, existing models mainly still work on the original atomic space, while our method works on the fundamentally different and more expressive latent space.

\section{Background}

\subsection{Problem Definition}
\label{ssec:background:define}

In this paper, we consider generative modeling of molecular geometries from scratch. Let $d$ be the dimension of node features, then each molecule is represented as point clouds $\gG= \langle \rvx, \rvh \rangle$, where $\rvx = (\rvx_1, \dots, \rvx_N )\in \mathbb{R}^{N \times 3}$ is the atom coordinates matrix and $\rvh = (\rvh_1, \dots, \rvh_N )\in \mathbb{R}^{N \times d}$ is the node feature matrix, such as atomic type and charges.
We consider the following two generation tasks:\\
\textbf{(I) Unconditional generation.} With a collection of molecules $\gG$, learn parameterized generative models $\ptheta(\gG)$ which can generate diverse and realistic molecules $\hat{\gG}$ in 3D.\\
\textbf{(II) Controllable generation.} With molecules $\gG$ labeled with certain properties $s$, learn conditional generation models $\ptheta(\gG | s)$ which can conduct controllable molecule generation given desired property value $s$.


\subsection{Equivariance}
\label{ssec:background:equivariance}

\textit{Equivariance} is ubiquitous for geometric systems such as molecules, where vector features like atomic forces or dipoles should transform accordingly \textit{w.r.t.} the coordinates~\citep{Thomas2018TensorFN, Weiler20183DSC, fuchs2020se3, batzner2021se}.
Formally, a function $\gF$ is defined as equivariant \textit{w.r.t} the action of a group $G$ if 
$
    \mathcal{F} \circ S_g (\rvx) = T_g \circ \mathcal{F} (\rvx), \forall g \in G
$
where $S_g, T_g$ are transformations for a group element $g$ \citep{serre1977linear}.
In this work, we consider the Special Euclidean group SE(3), \textit{i.e.}, the group of rotation and translation in 3D space, where transformations $T_g$ and $S_g$ can be represented by a translation $\vt$ and an orthogonal matrix rotation $\rmR$.

In molecules the features $\rvh$ are SE(3)-invariant while the coordinates will be affected\footnote{We follow the convention to use $\rmR\rvx$ to denote applying group actions $\rmR$ on $\rvx$, which formally is calculated as $\rvx \rmR^T$.} as $\rmR \rvx + \vt = (\rmR\rvx_1 + \vt, \dots, \rmR\rvx_N + \vt)$.
This requires our learned likelihood to be invariant to roto-translations. Such property has been shown important for improving the generalization capacity of 3D geometric modeling~\citep{satorras2021enflow,xu2022geodiff}.

\subsection{Diffusion Models for Non-geometric Domains}
\label{ssec:background:diffusion}

Diffusion models \citep{sohl2015deep,ho2020denoising} are latent variable models that model the data $\cx{0}$ as Markov chains $\cx{T} \cdots \cx{0}$, with intermediate variables sharing the same dimension. 
DMs can be described with two Markovian processes: a forward \textit{diffusion} process $q(\cx{1:T} \mid \cx{0}) = \prod_{t=1}^T q(\cx{t} \mid \cx{t-1})$ and a reverse \textit{denoising} process $\ptheta(\cx{0:T}) = p(\cx{T}) \prod_{t=1}^T \ptheta(\cx{t-1} \mid \cx{t})$. The forward process gradually adds Gaussian noise to data $\cx{t}$:
\begin{equation}
\label{eq:ddpm_diffusion}
    q(\cx{t} \mid \cx{t-1}) = \mathcal{N} (\cx{t} ; \sqrt{1-\beta_t} \cx{t-1}, \beta_t \mI),
\end{equation}
where the hyperparameter $\beta_{1:T}$ controls the amount of noise added at each timestep $t$. The $\beta_{1:T}$ are chosen such that samples $\cx{T}$ can approximately converge to standard Gaussians, \textit{i.e.}, $q(\cx{T}) \approx \mathcal{N} (0, \mI)$.
Typically, this forward process $q$ is predefined without trainable parameters.

The generation process of DMs is defined as learning a parameterized reverse \textit{denoising} process, which aims to incrementally denoise the noisy variables $\cx{T:1}$ to approximate clean data $\cx{0}$ in the target data distribution:
\begin{equation}
\label{eq:ddpm_denoising}
    \ptheta(\cx{t-1}\mid \cx{t}) = \mathcal{N}(\cx{t-1}; \vmu_\theta(\cx{t}, t), \rho_t^2 \mI),
\end{equation}
where the initial distribution $p(\cx{T})$ is defined as $\mathcal{N} (0, \mI)$. The means $\vmu_\theta$ typically are neural networks such as U-Nets for images or Transformers for text, and the variances $\rho_t$ typically are also predefined. 

As latent variable models, the forward process $q(\cx{1:T} | \cx{0})$ can be viewed as a fixed posterior, to which the reverse process $\ptheta(\cx{0:T})$ is trained to maximize the variational lower bound of the likelihood of the data
$
    \gL_\textit{vlb} =  \mathbb{E}_{q(\cx{1:T} | \cx{0})}  \Big[\log \frac{q(\cx{T} | \cx{0})}{\ptheta(\cx{T})} + \sum_{t=2}^T \log \frac{q(\cx{t-1} | \cx{0},\cx{t})} {\ptheta(\cx{t-1} | \cx{t}) } - \log \ptheta( \cx{0} | \cx{1})\Big].
$
However, directly optimizing this objective is known to suffer serious training instability \cite{nichol2021improved}.
Instead, \citet{song2019generative,ho2020denoising} suggest a simple surrogate objective up to irrelevant constant terms:
\begin{equation}
\label{eq:ddpm_loss}
    \gL_\textit{DM} = \mathbb{E}_{\cx{0},\vepsilon \sim \mathcal{N} (0, \mI), t} \big[ w(t) || \vepsilon
    - \vepsilon_\theta(\cx{t}, t)||^2 \big],
\end{equation}
where $\cx{t} = \alpha_t \cx{0} + \sigma_t \vepsilon$, with $\alpha_t=\sqrt{\prod_{s=1}^t(1-\beta_s)}$ and $\sigma_t=\sqrt{1-\alpha_t^2}$ are parameters from the tractable diffusion distributions $q(\rvx_t|\rvx_0)=\gN(\rvx_t;\alpha_t\rvx_0,\sigma_t^2\mI)$. $\vepsilon_\theta$ comes from the widely adopted parametrization of the means $\mathbf{\mu}_\theta(\rvx_t,t):=\tfrac{1}{\sqrt{1-\beta_t}}\big(\rvx_t-\tfrac{\beta_t}{\sqrt{1-\alpha_t^2}}\vepsilon_\theta(\rvx_t,t)\big)$. 
The reweighting terms are $w(t)=\frac{\beta_t^2}{2\rho_t^2(1-\beta_t)(1-\alpha_t^2)}$, while in practice simply setting it as $1$ often promotes the sampling quality. Intuitively, the model $\vepsilon_\theta$ is trained to predict the noise vector $\vepsilon$ to denoise diffused samples $\rvx_t$ at every step $t$ towards a cleaner one $\rvx_{t-1}$. After training, we can draw samples with $\vepsilon_\theta$ by the iterative ancestral sampling:
\begin{equation} 
\label{eq:ddpm_sampling}
    \rvx_{t-1}=\tfrac{1}{\sqrt{1-\beta_t}}(\rvx_t -\tfrac{\beta_t}{\sqrt{1-\alpha_t^2}}\vepsilon_\theta(\rvx_t,t)) +\rho_t \vepsilon,
\end{equation}
with $\vepsilon \sim \gN(\bm{0}, \mI)$. The sampling chain is initialized from Gaussian prior $\rvx_T \sim p(x_T)=\gN(\rvx_T;\bm{0}, \mI)$.

\section{Method}

In this section, we formally describe Geometric Latent Diffusion Models (\method). Our work is inspired by the recent success of stable (latent) diffusion models~\citep{rombach2022high}, but learning latent representations for the geometric domain is however challenging~\citep{winter2021auto}. We address these challenges by learning a faithful point-structured latent space with both invariant and equivariant variables, and elaborate on the design details of geometric autoencoding and latent diffusion in \cref{ssec:method:autoencoder} and \cref{ssec:method:latentdiff} respectively. Finally, we briefly summarize the simple training and sampling scheme in \cref{ssec:method:train}, and further discuss extensions for conditioning mechanisms in \cref{ssec:method:conditional}. A high-level schematic is provided in \cref{fig:framework}.


\subsection{Geometric Autoencoding}
\label{ssec:method:autoencoder}

We are interested in first compressing the geometries $\gG = \langle \rvx, \rvh \rangle \in \mathbb{R}^{N \times (3+d)}$ (see \cref{ssec:background:define} for details) into lower-dimensional latent space. We consider the classic autoencoder (AE) framework, where the encoder $\gE_\phi$ encodes $\gG$ into latent domain $\rvz = \gE_\phi (\rvx, \rvh)$ and the decoder $\gD_\xi$ learns to decode $\rvz$ back to data domain $\tilde{\rvx}, \tilde{\rvh} = \gD_\xi(\rvz)$. The whole framework can be trained by minimizing the reconstruction objective $\vd(\gD(\gE(\gG)), \gG)$, \textit{e.g.}, $L_p$ norms.

However, this classic autoencoding scheme is non-trivial in the geometric domain. Considering we follow SE(3) group in this paper (see \cref{ssec:background:equivariance}), the typical parameterization of latent space as invariant
scalar-valued features~\citep{kingma2013auto} is very challenging:
\begin{proposition} 
\label{prop:se3autoencoding}
\citep{winter2022unsupervised}
Learning autoencoding functions $\gE$ and $\gD$ to represent geometries $\gG$ in scalar-valued (i.e., invariant) latent space \textbf{necessarily} requires an additional \textbf{equivariant} function $\psi$ to store \textbf{suitable} group actions such that $\gD(\psi(\gG),\gE(\gG)) = T_{\psi(\gG)} \circ \hat{\gD}(\gE(\gG)) = \gG$.
\end{proposition}
The idea of this proposition is that Geometric AE requires an additional function $\psi$ to represent appropriate group actions for encoding, and align output and input positions for decoding, to solve the reconstruction task. 
We leave a more detailed explanation with examples in \cref{app:sec:prop-ae}. 
For euclidean groups SE(n), \citet{winter2022unsupervised} suggests implementing $\psi$ as equivariant ortho-normal vectors in the unit n-dimensional sphere $S^n$.

In our method, instead of separately representing and applying the equivariance with $\psi$, we propose to also incorporate equivariance into $\gE$ and $\gD$ by constructing latent features as point-structured variables $\rvz=\langle \rvz_\rx, \rvz_\rh \rangle \in \mathbb{R}^{N \times (3+k)}$, which holds $3$-d equivariant and $k$-d invariant latent features $\rvz_\rx$ and $\rvz_\rh$ for each node. This in practice can be implemented by parameterizing $\gE$ and $\gD$ with equivariant graph neural networks (EGNNs)~\citep{satorras2021en}, which extract both invariant and equivariant embeddings with the property:
\begin{equation}
\label{eq:equivariance:ae}
    \mathbf{R} \rvz_\rx + \vt, \rvz_\rh = \gE_\phi(\mathbf{R}\rvx + \vt, \rvh); \mathbf{R} \rvx + \vt, \rvh = \gD_\xi(\mathbf{R}  \rvz_\rx + \vt, \rvz_\rh),
\end{equation}
for all rotations $\mathbf{R}$ and translations $\vt$.
We provide parameterization details of EGNNs in \cref{app:sec:models}. The latent points $\rvz_\rx$ can perform the role of $\psi$ required in \cref{prop:se3autoencoding}, to align the orientation of outputs towards inputs. Furthermore, this point-wise latent space follows the inherent structure of geometries $\gG$, thereby achieving good reconstructions. 

Then the encoding and decoding processes can be formulated by $q_\phi (\rvz_\rx, \rvz_\rh | \rvx, \rvh) = \gN (\gE_\phi(\rvx, \rvh), \sigma_0 \mI)$ and $p_\xi (\rvx, \rvh | \rvz_\rx, \rvz_\rh) = \prod_{i=1}^N p_\xi (x_i, h_i | \rvz_\rx, \rvz_\rh)$ respectively. Following \citet{xu2022geodiff,hoogeboom2022equivariant} that linear subspaces with the center of gravity always being zero can induce translation-invariant distributions, we also define distributions of latent $\rvz_\rx$ and reconstructed $\rvx$ on the subspace that $\sum_i \rvz_{\rx,i} \text{ (or $\rvx_i$)} = 0$. The whole framework can be effectively optimized by:
\begin{equation}
\begin{aligned}
\label{eq:loss_ae}
    & \gL_\textit{AE} = \gL_\textit{recon} + \gL_\textit{reg}, \\
    & \gL_\textit{recon} = - \mathbb{E}_{q_\phi (\rvz_\rx, \rvz_\rh | \rvx, \rvh)} p_\xi (\rvx, \rvh | \rvz_\rx, \rvz_\rh),
\end{aligned}
\end{equation}
which is a reconstruction loss combined with a regularization term. The reconstruction loss in practice is calculated as $L_2$ norm or cross-entropy for continuous or discrete features. For the $\gL_\textit{reg}$ terms we experimented with two variants: \textit{KL-reg}~\citep{rombach2022high}, a slight Kullback-Leibler penalty of $q_\phi$ towards standard Gaussians similar to variational AE; and \textit{ES-reg}, an early-stop $q_\phi$ training strategy to avoid a scattered latent space. The regularization prevents latent embeddings from arbitrarily high variance and is thus more suitable for learning the latent DMs (LDMs).

\subsection{Geometric Latent Diffusion Models}
\label{ssec:method:latentdiff}

With the equivariant autoencoding functions $\gE_\phi$ and $\gD_\xi$, now we can represent structures $\gG$ using lower-dimensional latent variables $\rvz$
while still keeping geometric properties. 
Compared with the original atomic features which are high-dimensional with complicated data types and scales, the encoded latent space significantly benefits likelihood-based generative models since: (i) as described in \cref{ssec:method:autoencoder}, our proposed AEs can be viewed as \textit{regularized autoencoders}~\citep{Ghosh2020From}, where the latent space is more compact and smoothed, thereby improving DM's training; (ii) latent codes also enjoy lower dimensionality and benefit the generative modeling complexity, since DMs typically operate in the full dimension of inputs.

Existing latent generative models for images~\citep{vahdat2021score,esser2021taming} and texts~\citep{li2022diffusionlm} usually rely on typical autoregressive or diffusion models to model the scalar-valued latent space. 
By contrast, a fundamental challenge for our method is that the latent space $\rvz$ contains not only scalars (\textit{i.e,}, invariant features) $\rvz_\rh$ but also tensors (\textit{i.e,}, equivariant features) $\rvz_\rx$. This requires the distribution of latent DMs to satisfy the critical invariance:
\begin{equation}
    \ptheta(\rvz_\rx, \rvz_\rh) = \ptheta(\rmR \rvz_\rx, \rvz_\rh), \text{ $\forall$ $\rmR$}.
\end{equation}
\citet{xu2022geodiff} proved that this can be achieved if the initial distribution $p(\rvz_{\rx,T}, \rvz_{\rh,T})$ is invariant while the transitions $\ptheta(\rvz_{\rx, t-1}, \rvz_{\rh, t-1} | \rvz_{\rx,t}, \rvz_{\rh,t})$ are equivariant:
\begin{equation}
\begin{aligned}
    \ptheta(\rvz_{\rx, t-1}, \rvz_{\rh, t-1} & | \rvz_{\rx,t}, \rvz_{\rh,t}) = \\
    & \ptheta(\rmR \rvz_{\rx, t-1}, \rvz_{\rh, t-1} | \rmR \rvz_{\rx,t}, \rvz_{\rh,t}), \text{ $\forall$ $\rmR$}.
\end{aligned}
\end{equation}
\citet{xu2022geodiff,hoogeboom2022equivariant} further show that this can be realized by implementing the denoising dynamics $\vepsilon_\theta$ with equivariant networks such that:
\begin{equation}
\label{eq:equivariance:ldm}
    \rmR \rvz_{\rx, t-1} + \vt, \rvz_{\rh, t-1} = \vepsilon_\theta(\rmR \rvz_{\rx,t} + \vt, \rvz_{\rh,t}, t), \text{ $\forall$ $\rmR$ and $\vt$}.
\end{equation}
which in practice we parameterize as time-conditional EGNNs. 
More model details are also provided in \cref{app:sec:models}. Similar to the encoding posterior, in order to keep translation invariance, all the intermediate states $\rvz_{\rx,t}, \rvz_{\rh,t}$ are also required to lie on the subspace by $\sum_i \rvz_{\rx,t,i} = 0$ by moving the center of gravity. Analogous to \cref{eq:ddpm_loss}, now we can train the model by:
\begin{equation}
\label{eq:loss_ldm}
    \gL_\textit{LDM} = \mathbb{E}_{\gE(\gG),\vepsilon \sim \mathcal{N} (0, \mI), t} \big[ w(t) || \vepsilon
    - \vepsilon_\theta(\rvz_{\rx,t}, \rvz_{\rh,t}, t)||^2 \big],
\end{equation}
with $w(t)$ simply set as $1$ for all steps $t$.

\begin{algorithm}[!t]
   \caption{Training  Algorithm of \method}
   \label{alg:training}
\begin{algorithmic}[1]
\STATE{\bf Input:} geometric data $\gG=\langle \rvx, \rvh \rangle$
\STATE{\bf Initial:} encoder network $\gE_\phi$, decoder network $\gD_\xi$, denoising network $\vepsilon_\theta$
\STATE{\bf First Stage: Autoencoder Training}
    \WHILE{$\phi,\xi$ have not converged}
    \STATE $\vmu_\rx, \vmu_\rh \leftarrow \gE_\phi(\rvx, \rvh)$ \hfill\COMMENT{Encoding}
    \STATE $\vepsilon \sim \gN(\bm{0}, \mI)$
    \STATE Subtract center of gravity from $\vepsilon_\rx$ in $\vepsilon = [\vepsilon_\rx, \vepsilon_\rh]$
    \STATE $\rvz_\rx, \rvz_\rh \leftarrow \vepsilon \odot \sigma_0 + \vmu $ \hfill\COMMENT{Reparameterization}
    \STATE $\Tilde{\rvx}, \tilde{\rvh} \leftarrow \gD_\xi (\rvz_\rx, \rvz_\rh)$ \hfill\COMMENT{Decoding}
    \STATE $\gL_\textit{AE} = \operatorname{reconstrcution}([\Tilde{\rvx}, \tilde{\rvh}],[\rvx, \rvh]) + \gL_\textit{reg}$
    \STATE $\phi, \xi \leftarrow \operatorname{optimizer}(\gL_\textit{AE};\phi,\xi)$
\ENDWHILE
\STATE{\bf Second Stage: Latent Diffusion Models Training}
\STATE Fix encoder parameters $\phi$
\WHILE{$\theta$ have not converged}
    \STATE $\rvz_{\rx,0}, \rvz_{\rh,0} \sim q_\phi (\rvz_\rx, \rvz_\rh | \rvx, \rvh)$ \hfill \COMMENT{As lines 5-8}
    \STATE $t \sim \rmU(0,T)$, $\vepsilon \sim \gN(\bm{0}, \mI)$
    \STATE Subtract center of gravity from $\vepsilon_\rx$ in $\vepsilon = [\vepsilon_\rx, \vepsilon_\rh]$
    \STATE $\rvz_{\rx,t}, \rvz_{\rh,t} = \alpha_t [\rvz_{\rx,0}, \rvz_{\rh,0}] + \sigma_t \vepsilon$
    \STATE $\gL_\textit{LDM} = || \vepsilon
    - \vepsilon_\theta(\rvz_{\rx,t}, \rvz_{\rh,t}, t)||^2 $
    \STATE $\theta \leftarrow \operatorname{optimizer}(\gL_\textit{LDM}; \theta)$
\ENDWHILE
\STATE \bf{return} $\gE_\phi$, $\gD_\xi$, $\vepsilon_\theta$
\end{algorithmic}
\end{algorithm}

\textbf{Theoretical analysis.} The combined objective for the whole framework, \textit{i.e.}, $\gL_\textit{AE} + \gL_\textit{LDM}$, appears similar to the standard VAE objective with an additional regularization. We make the formal justification that considering neglecting the minor $\gL_{reg}$ term, $\gL = \gL_\textit{recon} + \gL_\textit{LDM}$ is theoretically an SE(3)-invariant variational lower bound of log-likelihood:
\begin{theorem}
\label{theorem:elbo}
(informal) Let $\gL \mathrel{\mathop:}= \gL_\textit{recon} + \gL_\textit{LDM}$. With certain weights $w(t)$, $\gL$ is an SE(3)-invariant variational lower bound to the log-likelihood, \textit{i.e.}, for any geometries $\langle \rvx, \rvh \rangle$, we have:
\begin{align}
    & \gL(\rvx, \rvh) \geq - \mathbb{E}_{p_\textit{data}} [\log p_{\theta,\xi} (\rvx, \rvh)],\text{ and} \nonumber\\
    & \gL(\rvx, \rvh) = \gL(\rmR \rvx  + \vt, \rvh),\text{ $\forall$ rotation $\rmR$ and translation $\vt$}, \nonumber
\end{align}
where $p_{\theta,\xi} (\rvx, \rvh) = \mathbb{E}_ {p_\theta (\rvz_\rx, \rvz_\rh)} p_\xi(\rvx, \rvh | \rvz_\rx, \rvz_\rh) $ is the marginal distribution of $\langle \rvx, \rvh \rangle$ under \method model.
\end{theorem}
Furthermore, for the induced marginal distribution $p_{\theta,\xi} (\rvx, \rvh)$, we also hold the equivariance property that:
\begin{proposition}
\label{prop:inv-likelihood}
    With decoders and latent DMs defined with equivariant distributions, the marginal $p_{\theta,\xi} (\rvx, \rvh) = \mathbb{E}_ {p_\theta (\rvz_\rx, \rvz_\rh)} p_\xi(\rvx, \rvh | \rvz_\rx, \rvz_\rh)$ is an SE(3)-invariant distribution.
\end{proposition}
These theoretical analysis suggest that \method is parameterized and optimized in an SE(3)-invariant fashion, which is a critical inductive bias for geometric generative models~\citep{satorras2021enflow,xu2022geodiff} and provides explanations as to why our framework can achieve better 3D geometries generation quality. We provide the full statements and proofs in \cref{app:sec:proof}

\begin{figure*}[!t]
    \centering
    \includegraphics[width=1.0\linewidth]{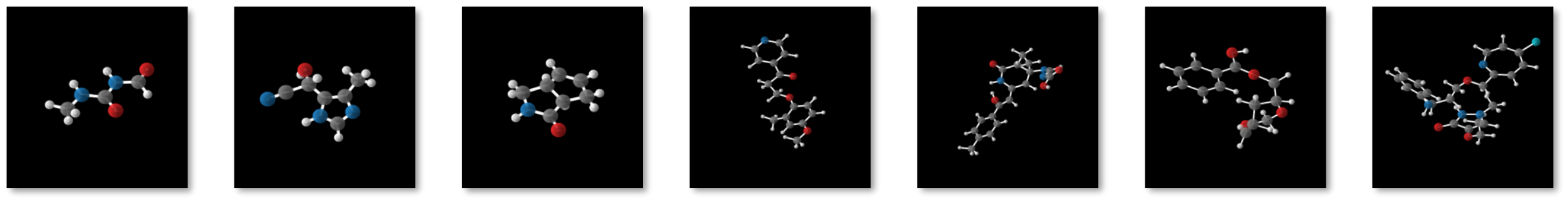}
    \vspace{-20pt}
    \caption{Molecules generated by \method trained on QM9 (left three) and DRUG (right four).}
    \label{fig:vis-unconditional}
\end{figure*}

\subsection{Training and Sampling}
\label{ssec:method:train}

With the proposed formulation and practical parameterization, we now present the training and sampling schemes for \method. While objectives for training Geometric AEs and LDMs are already defined in \cref{eq:loss_ae,eq:loss_ldm}, it is still unclear whether the two components should be trained one by one, or optimized simultaneously by backpropagation through reparameterizing~\citep{kingma2013auto}.
Previous work about latent DMs for image generation~\citep{sinha2021d2c,rombach2022high} shows that the two-stage training strategy usually leads to better performance, and we notice similar phenomena in our experiments. 
This means we first train AE with regularization, and then train the latent DMs on the latent embeddings encoded by the pre-trained encoder.
A formal description of the training process is provided in \cref{alg:training}.

With \method we can formally define a residual generative distribution $p_{\theta,\xi}(\rvx, \rvh, \rvz_\rx, \rvz_\rh) = p_\theta(\rvz_\rx, \rvz_\rh) p_\xi (\rvx, \rvh | \rvz_\rx, \rvz_\rh)$, where $p_\theta$ refers to the latent DM modeling the point-structured latent codes, and $p_\xi$ denotes the decoder.
We can generate molecular structures by first sampling equivariant latent embeddings from $\ptheta$ and then translating them back to the original geometric space with $p_\xi$. The pseudo-code of the sampling procedure is provided in \cref{alg:sampling}.

\begin{algorithm}[!t]
   \caption{Sampling Algorithm of \method}
   \label{alg:sampling}
\begin{algorithmic}[1]
\STATE{\bf Input:} decoder network $\gD_\xi$, denoising network $\vepsilon_\theta$
\STATE $\rvz_{\rx,T}, \rvz_{\rh,T} \sim \gN(\mathbf{0}, \mI)$
\FOR{$t$ in $T,\, T-1, \cdots, 1$}
    \STATE $\vepsilon \sim \mathcal{N}(\mathbf{0}, \mI)$ \hfill \COMMENT{Latent Denoising Loop}
    \STATE Subtract center of gravity from $\vepsilon_\rx$ in $\vepsilon = [\vepsilon_\rx, \vepsilon_\rh]$
    \STATE $\rvz_{t-1}=\tfrac{1}{\sqrt{1-\beta_t}}(\rvz_t -\tfrac{\beta_t}{\sqrt{1-\alpha_t^2}}\vepsilon_\theta(\rvz_t,t)) +\rho_t \vepsilon$
\ENDFOR
\STATE $\rvx, \rvh \sim p_\xi (\rvx, \rvh | \rvz_{\rx,0}, \rvz_{\rh,0})$ \hfill \COMMENT{Decoding}
\STATE \textbf{return} $\rvx, \rvh$
\end{algorithmic}
\end{algorithm}

For the number of nodes $N$, in the above sections, we assume it to be predefined for each data point. In practice, we need to sample different numbers $N$ for generating molecules of different sizes. We follow the common practice~\citep{satorras2021enflow} to first count the distribution $p(N)$ of molecular sizes on the training set. Then for generation, we can first sample $N \sim p(N)$ and then generate latent variables and node features in size $N$.

\subsection{Controllable Generation}
\label{ssec:method:conditional}

Similar to other generative models~\citep{kingma2013auto,van2016pixel}, DMs are also capable of controllable generation with given conditions $s$, by modeling conditional distributions $p(\rvz|s)$. This in DMs can be implemented with conditional denoising networks $\vepsilon_\theta(\rvz, t, s)$, with the critical difference that it takes additional inputs $s$.
In the molecular domain, desired conditions $s$ typically are chemical properties, which are much lower-dimensional than the text prompts for image generations~\citep{rombach2022high,ramesh2022hierarchical}. Therefore, instead of sophisticated cross-attention mechanisms used in text-guided image generation, we follow \citet{hoogeboom2022equivariant} and simply parameterize the conditioning by concatenating $s$ to node features. Besides, as a whole framework, we also adopt similar concatenation methods for the encoder and decoder, \textit{i.e.}, $\gE_\phi(\rvx, \rvh, s)$ and $\gD_\xi(\rvz_\rx, \rvz_\rh, s)$, to further shift the latent codes towards data distribution with desired properties $s$.

\begin{table*}[!t]
\centering
\caption{Results of atom stability, molecule stability, validity, and validity$\times$uniqueness. A higher number indicates a better generation quality. Metrics are calculated with 10000 samples generated from each model. On QM9, we run the evaluation for 3 times and report the derivation. Note that, for DRUG dataset, molecule stability and uniqueness metric are omitted since they are nearly $0\%$ and $100\%$ respectively for all the methods. Compared with previous methods, the latent space with both invariant and equivariant variables enables \method to achieve up to 7\% improvement for the validity of large molecule generation.
}
\label{tab:qm9_results}
\resizebox{\textwidth}{!}{
\begin{threeparttable}
\begin{tabular}{l | c c c c | c c}
    \toprule[1.0pt]
    & \multicolumn{4}{c|}{\shortstack[c]{\textbf{QM9}}} & \multicolumn{2}{c}{\shortstack[c]{\textbf{DRUG}}} \\
    \# Metrics & Atom Sta (\%) & Mol Sta (\%) & Valid (\%) & Valid \& Unique (\%) & Atom Sta (\%) & Valid (\%) \\
    \midrule[0.8pt]
    Data &  99.0 & 95.2 & 97.7 & 97.7 & 86.5 & 99.9 \\
    \midrule
    {ENF} &  85.0 & 4.9 & 40.2 & 39.4 & - & - \\
    {G-Schnet} & 95.7 & 68.1 & 85.5 & 80.3 & - & - \\
    GDM & 97.0 & 63.2 & - & - & 75.0 & 90.8 \\ 
    GDM-\textsc{aug} & 97.6 & 71.6 & 90.4 & 89.5 & 77.7 & 91.8\\ 
    EDM & 98.7 & 82.0 & 91.9 & 90.7 & 81.3 & 92.6 \\
    EDM-Bridge & 98.8 & 84.6 & 92.0* & 90.7 & 82.4 & 92.8* \\
    \midrule[0.3pt]
    \textbf{\textsc{GraphLDM}} & 97.2 & 70.5 & 83.6 & 82.7 & 76.2 & 97.2 \\
    \textbf{\textsc{GraphLDM-aug}} & 97.9 & 78.7 & 90.5 & 89.5 & 79.6 & 98.0 \\
    \rowcolor{lightgray} \textbf{\method} & \textbf{98.9} $\pm$ 0.1 & \textbf{89.4} $\pm$ 0.5 & \textbf{93.8} $\pm$ 0.4 & \textbf{92.7} $\pm$ 0.5 & \textbf{84.4} & \textbf{99.3} \\
    \bottomrule[1.0pt]
\end{tabular}
\begin{tablenotes}
\small
\item *Results obtained by our own experiments. Other results are borrowed from recent studies \cite{hoogeboom2022equivariant,wu2022diffusionbased}.
\end{tablenotes}
\end{threeparttable}
}
\vspace{-10pt}
\end{table*}

\section{Experiments}

In this section, we justify the advantages of \method with comprehensive experiments.
We first introduce our experimental setup in \cref{ssec:exp-setup}.
Then we report and analyze the evaluation results in \cref{ssec:exp-molgen} and \cref{ssec:exp-condition}, for unconditional and conditional generation respectively. We also provide further ablation studies in \cref{app:sec:exp-ablation} to investigate the effect of several model designs.
We leave more implementation details in \cref{app:sec:exp-details}.

\subsection{Experiment Setup}
\label{ssec:exp-setup}

\textbf{Evaluation Task.}
Following previous works on molecule generation in 3D~\citep{gebauer2019symmetry,luo2021autoregressive,satorras2021enflow,hoogeboom2022equivariant,wu2022diffusionbased}, we evaluate \method by comparing with the state-of-the-art approaches on three comprehensive tasks. \textit{Molecular Modeling and Generation} measures the model’s capacity to learn the molecular data distribution and generate chemically valid and structurally diverse molecules. 
\textit{Controllable Molecule Generation} concentrates on generating target molecules with desired chemical properties. For this task, we retrain the conditional version \method on molecular data with corresponding property labels.

\textbf{Datasets.} We first adopt \textit{QM9} dataset~\citep{ramakrishnan2014quantum} for both unconditional and conditional molecule generation. QM9 is one of the most widely-used datasets for molecular machine learning research, which has also been adopted in previous 3D molecule generation studies~\citep{gebauer2019symmetry,gebauer2021inverse}. QM9 contains 3D structures together with several quantum properties for 130k small molecules, limited to 9 heavy atoms (29 atoms including hydrogens). 
Following~\citep{anderson2019cormorant}, we split the train, validation, and test partitions, with 100K, 18K, and 13K samples. For the molecule generation task, we also test \method on the \textit{GEOM-DRUG} (Geometric Ensemble Of Molecules) dataset. The DRUG dataset consists of much larger organic compounds, with up to 181 atoms and 44.2 atoms on average, in 5 different  atom types. It covers 37 million molecular conformations for around 450,000 molecules, labeled with energy and statistical weight. 
We follow the common practice~\citep{hoogeboom2022equivariant} to select the 30 lowest energy conformations of each molecule for training.


\subsection{Molecular Modeling and Generation}
\label{ssec:exp-molgen}

\textbf{Evaluation Metrics.} We measure model performances by evaluating the chemical feasibility of generated molecules, indicating whether the model can learn chemical rules from data. Given molecular geometries, we first predict bond types (single, double, triple, or none) by pair-wise atomic distances and atom types. Then we calculate the \textit{atom stability} and \textit{molecule stability} of the predicted molecular graph. The first metric captures the proportion of atoms that have the right valency, while the latter is the proportion of generated molecules for which all atoms are stable. In addition,  We report \textit{validity} and \textit{uniqueness} metrics, which are the percentages of valid (measured by \textsc{RDKit}) and unique molecules among all the generated compounds.

\textbf{Baselines.}
We compare \method to several competitive baseline models. \textit{G-Schnet}~\citep{gebauer2019symmetry} and Equivariant Normalizing Flows (\textit{ENF})~\citep{satorras2021enflow} are previous equivariant generative models for molecules, based on autoregressive and flow-based models respectively. Equivariant Graph Diffusion Models (\textit{EDM}) with its non-equivariant variant (\textit{GDM})~\citep{hoogeboom2022equivariant} are recent progress on diffusion models for molecule generation. Most recently, \citet{wu2022diffusionbased} proposed an improved version of EDM (\textit{EDM-Bridge}), which further boosts the performance with well-designed informative prior bridges. To yield a fair comparison, all the baseline models use the same parameterization and training configurations as described in \cref{ssec:exp-setup}.


\begin{figure*}[!t]
    \centering
    \includegraphics[width=1.0\linewidth]{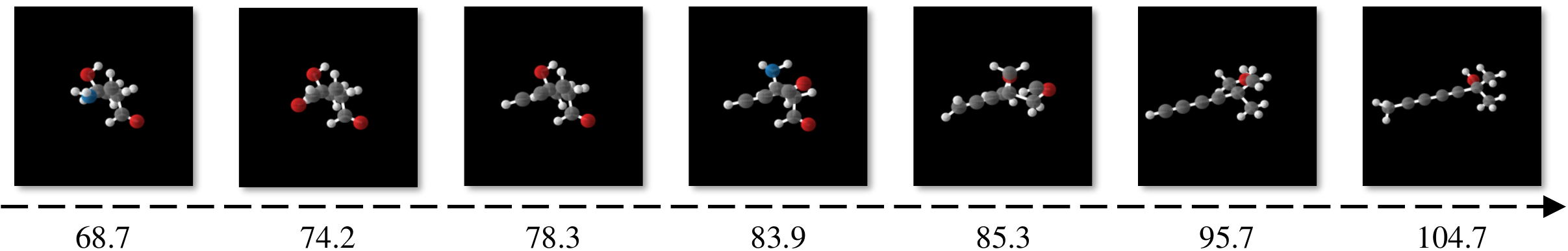}
    \vspace{-20pt}
    \caption{Molecules generated by conditional \method. We conduct controllable generation with interpolation among different Polarizability $\alpha$ values with the same reparametrization noise $\vepsilon$. The given $\alpha$ values are provided at the bottom.}
    \label{fig:vis-conditional}
    \vspace{-10pt}
\end{figure*}

\textbf{Results and Analysis.}
We generate $10,000$ samples from each method to calculate the above metrics, and the results are reported in \cref{tab:qm9_results}. As shown in the table, \method outperforms competitive baseline methods on all metrics with an obvious margin. 
It is worth noticing that, for the DRUG dataset, even ground-truth molecules have $86.5\%$ atom-level and nearly $0\%$ molecule-level stability. This is because the DRUG molecules contain larger and more complex structures, creating errors during bond type prediction based on pair-wise atom types and distances. Furthermore, as DRUG contains many more molecules with diverse compositions, we also observe that \textit{unique} metric is almost $100\%$ for all methods. Therefore, we omit the \textit{molecule stability} and \textit{unique} metrics for the DRUG dataset. Overall, the superior performance demonstrates \method's higher capacity to model the molecular distribution and generate chemically realistic molecular geometries.
We provide visualization of randomly generated molecules in \cref{fig:vis-unconditional}, and leave more visualizations in \cref{app:sec:vis}.

\textbf{Ablation Study.}
Furthermore, to verify the benefits of incorporating equivariant latent features, we conduct ablation studies with only invariant variables in the latent space, called Graph Latent Diffusion Models (\textsc{GraphLDM}). 
We run \textsc{GraphLDM} with the same configuration as our method, except that all modules (\textit{i.e.}, encoder, decoder, and latent diffusion models) are instead equipped with typical non-equivariant graph networks. 
We also follow \citet{hoogeboom2022equivariant} to test \textsc{GDM-aug} and \textsc{GraphLDM-aug}, where models are trained with data augmented by random rotations. \cref{tab:qm9_results} shows the empirical improvement of \method over these ablation settings, which verifies the effectiveness of our latent equivariance design.

\subsection{Controllable Molecule Generation}
\label{ssec:exp-condition}

\begin{table}[!t] 
\setlength{\tabcolsep}{2pt}
\centering
\caption{Mean Absolute Error for molecular property prediction. A lower number indicates a better controllable generation result. Results are predicted by a pretrained EGNN classifier $\omega$ on molecular samples extracted from individual methods.
}
\label{tab:conditional_results}
\resizebox{\linewidth}{!}{
\begin{threeparttable}
    \begin{tabular}{l | c c c c c c}
    \toprule
    Property & $\alpha$& $\Delta \varepsilon$ & $\varepsilon_{\mathrm{HOMO}}$ & $\varepsilon_{\mathrm{LUMO}}$ & $\mu$ & $C_v$\\
    Units & Bohr$^3$ & meV & meV & meV & D & $\frac{\text{cal}}{\text{mol}}$K  \\
    \midrule
    QM9* & 0.10  & 64 & 39 & 36 & 0.043 & 0.040  \\
    \midrule
    Random* & 9.01  &  1470 & 645 & 1457  & 1.616  & 6.857   \\
    $N_\text{atoms}$ & 3.86  & 866 & 426 & 813 & 1.053  &  1.971 \\
    EDM             & 2.76  & 655 & 356 & 584 & 1.111 & 1.101 \\
    \rowcolor{lightgray} 
    \textbf{\method} & 2.37 & 587 & 340 & 522 & 1.108 & 1.025 \\
    \bottomrule
\end{tabular}
\begin{tablenotes}
\small
\item *The results of \textit{QM9} and \textit{Random} can be viewed as lower and upper bounds of MAE on all properties.
\end{tablenotes}
\end{threeparttable}
}
\vspace{-10pt}
\end{table}

\textbf{Evaluation Metrics.}
In this task, we aim to conduct controllable molecule generation with the given desired properties. 
This can be useful in realistic settings of material and drug design where we are interested in discovering molecules with specific property preferences.
We test our conditional version of \method on QM9 with 6 properties: polarizability $\alpha$, orbital energies $\varepsilon_{\mathrm{HOMO}}$, $\varepsilon_{\mathrm{LUMO}}$ and their gap $\Delta \varepsilon$, Dipole moment $\mu$, and heat capacity $C_v$.
For evaluating the model's capacity to conduct property-conditioned generation, we follow \citet{satorras2021enflow} to first split the QM9 training set into two halves with $50K$ samples in each.
Then we train a property prediction network $\omega$ on the first half, and train conditional models on the second half. 
Afterward, given a range of property values $s$, we conditionally draw samples from the generative models and then use $\omega$ to calculate their property values as $\hat{s}$.
We report the \textit{Mean Absolute Error (MAE)} between $s$ and $\hat{s}$ to measure whether generated molecules are close to their conditioned property.
We also test the MAE of directly running $\omega$ on the second half QM9, named
\textit{QM9} in \cref{tab:conditional_results}, which measures the bias of $\omega$.
A smaller  gap with \textit{QM9} numbers indicates a better property-conditioning performance.

\textbf{Baselines.} 
We incorporate existing EDM as our baseline model. In addition, we follow \citet{hoogeboom2022equivariant} to also list two baselines agnostic to ground-truth property $s$, named \textit{Random} and $N_\textit{atoms}$. 
\textit{Random} means we simply do random shuffling of the property labels in the dataset and then evaluate $\omega$ on it. This operation removes any relation between molecule and property, which can be viewed as an upper bound of \textit{MAE} metric.
$N_\text{atoms}$ predicts the molecular properties by only using the number of atoms in the molecule. 
The improvement over \textit{Random} can verify the method is able to incorporate conditional property information into the generated molecules.
And overcoming $N_\text{atoms}$ further indicates the model can incorporate conditioning into molecular structures beyond the number of atoms.

\textbf{Results and Analysis.}
We first provide a visualization of controlled molecule generation by \method in \cref{fig:vis-conditional}, as qualitative assessments. We interpolate the conditioning property with different Polarizability values $\alpha$ while keeping the reparameterization noise $\vepsilon$ fixed. Polarizability refers to the tendency of matter, when subjected to an electric field, to acquire an electric dipole moment in proportion to that applied field. Typically, less isometrically molecular geometries lead to larger $\alpha$ values. This is consistent with our observed phenomenon in \cref{fig:vis-conditional}.

We report the numerical results in \cref{tab:conditional_results}. As shown in the table, \method significantly outperforms baseline models, including the previous diffusion model running on atomic features (EDM), on all the property metrics. 
The results demonstrate that by modeling in the latent space, \method acquired a higher capacity to incorporate given property information into the generation process.

\section{Conclusion and Future Work}
We presented \method, a novel latent diffusion model for molecular geometry generation.
While current models operate directly on high-dimensional, multi-modal atom features, \method overcomes their limitations by learning diffusion models over a continuous, lower-dimensional latent space. 
By building point-structured latent codes with both invariant scalars and equivariant tensors, \method is able to effectively learn latent representations while maintaining roto-translational equivariance. 
Experimental results demonstrate its significantly better capacity for modeling chemically realistic molecules. 
For future work, as a general and principled framework, \method can be extended to various 3D geometric generation applications, \textit{e.g.}, apply \method in more realistic drug discovery scenarios with given protein targets, or scale up \method for more challenging 3D geometries such as peptides and proteins.


\section*{Acknowledgements}
We thank Tailin Wu, Aaron Lou, Xiang Lisa Li, and Kexin Huang for discussions and for providing feedback on our manuscript.
We also gratefully acknowledge the support of
DARPA under Nos. HR00112190039 (TAMI), N660011924033 (MCS);
ARO under Nos. W911NF-16-1-0342 (MURI), W911NF-16-1-0171 (DURIP);
NSF under Nos. OAC-1835598 (CINES), OAC-1934578 (HDR), CCF-1918940 (Expeditions), 
NIH under No. 3U54HG010426-04S1 (HuBMAP),
Stanford Data Science Initiative, 
Wu Tsai Neurosciences Institute,
Amazon, Docomo, GSK, Hitachi, Intel, JPMorgan Chase, Juniper Networks, KDDI, NEC, and Toshiba.
We also gratefully acknowledge the support of NSF (\#1651565), ARO (W911NF-21-1-0125), ONR (N00014-23-1-2159), CZ Biohub, Stanford HAI.
We also gratefully acknowledge the support of Novo Nordisk A/S.
Minkai Xu thanks the generous support of Sequoia Capital Stanford Graduate Fellowship.


\bibliography{icml2023}
\bibliographystyle{icml2023}

\newpage
\appendix
\onecolumn



\section{Explanation of \cref{prop:se3autoencoding}}
\label{app:sec:prop-ae}

We first explain the intuition behind the theoretical justification of \cref{prop:se3autoencoding} by an example here. Considering given a input geometry $\gG = \langle \rvh, \rvx \rangle$, the encoder $\gE$ and decoder $\gD$, such that $\gG = \gD(\gE(\gG))$. Then we can transform $\gG$ by an action $g$ from SE(3)-group to $\hat{\gG} = T_g \gG = \langle \rvh, \rmR \rvx + \vt \rangle$ and input it into the autoencoders. Since the encoding function is invariant, we have $\gE(\gG) = \gE(\hat{\gG})$, and thus the reconstructed geometry however will still be $\gG = \gD(\gE(\hat{\gG}))$ instead of $\hat{\gG}$. This is problematic because we couldn't calculate the reconstruction error based on $\gG$ and $\hat{\gG}$, and a natural solution is that we need an additional function $\psi$ to extract the group action $g$. Then after decoding, we can apply the group action on generated $\gG$ to recover $\hat{\gG}$, thereby solving the problem

Formally, the explanation is that all elements can be expressed in terms of coordinates with respect to a given basis. So we should consider a canonical basis for all orbits, and learn the equivariant function $\psi$ to indicate to which orbit elements are decoded as “canonical”. For detailed theoretical analysis, we refer readers to \citet{winter2022unsupervised}.

\section{Formal Statements and Proofs}
\label{app:sec:proof}

\subsection{Relationship to SE(3)-invariant Likelihood: \cref{theorem:elbo}}

First, recall the informal theorem we provide in \cref{ssec:method:latentdiff}, which builds the connection between \method's objective and SE(3)-invariant maximum likelihood:
\begingroup
\def\thetheorem{\ref{theorem:elbo}}
\begin{theorem}
    (informal) Let $\gL \mathrel{\mathop:}= \gL_\textit{recon} + \gL_\textit{LDM}$. With certain weights $w(t)$, $\gL$ is an SE(3)-invariant variational lower bound to the log-likelihood, \textit{i.e.}, for any geometries $\langle \rvx, \rvh \rangle$, we have:
    \begin{align}
        & \gL(\rvx, \rvh) \geq - \mathbb{E}_{p_\textit{data}} [\log p_{\theta,\xi} (\rvx, \rvh)],\text{ and}\\
        & \gL(\rvx, \rvh) = \gL(\rmR \rvx  + \vt, \rvh),\text{ $\forall$ rotation $\rmR$ and translation $\vt$},
    \end{align}
    where $p_{\theta,\xi} (\rvz_\rx, \rvz_\rh) = \mathbb{E}_ {p_\theta (\rvz_\rx, \rvz_\rh)} p_\xi(\rvx, \rvh | \rvz_\rx, \rvz_\rh) $ is the marginal distribution of $\langle \rvx, \rvh \rangle$ under \method model.
\end{theorem}
\addtocounter{theorem}{-1}
\endgroup
Before providing the proof, we first present a formal version of the theorem:
\begin{theorem}(formal)
For predefined valid $\{\beta_i\}_{i=0}^{T}$, $\{\alpha_i\}_{i=0}^{T}$, and $\{\rho_i\}_{i=0}^{T}$, let $w(t)$ satisfy:
\begin{equation}
   w(t)=\frac{\beta_t^2}{2\rho_t^2(1-\beta_t)(1-\alpha_t^2)},\, \forall t \in [1, \cdots, T], \quad \text{and} \quad w(0) = -1.
\end{equation}
Let $\gL(\rvx, \rvh; \theta, \phi, \xi) \mathrel{\mathop:}= \gL_\textit{recon}(\rvx, \rvh; \phi, \xi) + \gL_\textit{LDM}(\rvz_\rx, \rvz_\rh; \theta)$. Then given the geometries $\langle \rvx, \rvh \rangle \in \R^{N \times (3+d)}$, we have:
\begin{align}
    &\label{app:eq:elbo} \gL(\rvx, \rvh) \geq - \mathbb{E}_{p_\textit{data}} [\log p_{\theta,\xi} (\rvx, \rvh)],\text{ and}\\
    &\label{app:eq:loss-invaraince} \gL(\rvx, \rvh) = \gL(\rmR \rvx  + \vt, \rvh),\text{ $\forall$ rotation $\rmR$ and translation $\vt$},
\end{align}
$p_{\theta,\xi} (\rvz_\rx, \rvz_\rh) = \mathbb{E}_ {p_\theta (\rvz_\rx, \rvz_\rh)} p_\xi(\rvx, \rvh | \rvz_\rx, \rvz_\rh) $ is the marginal distribution of $\langle \rvx, \rvh \rangle$ under \method model.
\end{theorem}
As shown in the theorem, the conclusion is composed of two statements, \textit{i.e.}, \cref{app:eq:elbo} and \cref{app:eq:loss-invaraince}. The first equation states that $\gL$ is a variational lower bound of the log-likelihood, and the second equation shows that the objective $\gL$ is further SE(3)-invariant, \textit{i.e.}, invariant to any rotational and translational transformations. Here, we provide the full proofs of the two statements separately. We first present the proof for \cref{app:eq:elbo}:
\begin{proof}[Proof of \cref{theorem:elbo} (\cref{app:eq:elbo})]
For analyzing the variational lower bound in \cref{app:eq:elbo}, we don't need to consider the different geometric properties of $\rvx$ and $\rvh$. Therefore, in this part, we use $\gG$ to denote $\langle \rvz_\rx, \rvz_\rh \rangle$, and $\rvz_\gG$ to denote $\langle \rvz_\rx, \rvz_\rh \rangle$. Besides, we interchangeably use $\rvz_\gG$ and $\rvz_\gG^{(0)}$ to denote the ``clean" latent variables at timestep $0$, and use $\rvz_\gG^{(T)}$ to denote the ``noisy" latent variables at timestep $T$. Then we have that:
\begin{equation}
\begin{aligned}
    \bb{E}_{\pdata(\gG)}[\log p_{\theta,\xi}(\gG)] & = \bb{E}_{\pdata(\gG)}\left[\log \int_{\rvz_\gG} p_\xi(\rvx | \rvz_\gG) \ptheta(\rvz_\gG)\right] \\
    & \geq \bb{E}_{\pdata(\gG), \qphi(\rvz_\gG|\gG)}[\log p_\xi(\gG | \rvz_\gG) + \log \ptheta(\rvz_\gG) - \log \qphi(\rvz_\gG | \gG)] &\text{Jensen's inequality} \\
     & = \bb{E}_{\pdata(\gG), \qphi(\rvz_\gG | \gG)}[\log p_\xi(\gG | \rvz_\gG)] - \KL(\qphi(\rvz_\gG | \gG) \Vert \ptheta(\rvz_\gG)). &\text{KL divergence}
\end{aligned}
\end{equation}
Compared with the objective $\gL$ for \method that:
\begin{equation}
\begin{aligned}
    \gL(\gG; \theta, \phi, \xi) & \mathrel{\mathop:}= \gL_\textit{recon}(\gG; \phi, \xi) + \gL_\textit{LDM}(\rvz_\gG; \theta) \\
    & = \bb{E}_{\pdata(\gG), \qphi(\rvz_\gG | \gG)}[\log p_\xi(\gG | \rvz_\gG)] + \gL_\textit{LDM}(\rvz_\gG; \theta),
\end{aligned}
\end{equation}
it is clear that we can complete the proof if we have:
\begin{equation}
\begin{aligned}
    \gL_\textit{LDM}(\rvz_\gG; \theta) & \geq \KL( \qphi(\rvz_\gG | \gG) \Vert \ptheta(\rvz_\gG)) \\ 
    & = - H(\qphi(\rvz_\gG | \gG)) - \bb{E}_{\qphi(\rvz_\gG | \gG)}[\log p_\theta(\rvz_\gG)]
\end{aligned}
\end{equation}
or since the Shannon entropy term $H(\qphi(\rvz_\gG | \gG))$ is never negative, we can equivalently prove:
\begin{align}
    \gL_\textit{LDM}(\rvz_\gG; \theta) \geq - \bb{E}_{\qphi(\rvz_\gG | \gG)}[\log p_\theta(\rvz_\gG)]
\end{align}
Now we prove the inequality by analyzing the right side of the inequality. We first apply variational inference with an inference model $q(\rvz_\gG^{({1:T})}|\rvz_\gG^{(0)})$. Note that, now we change the notation of ``clean" latent variable from $\rvz_\gG$ to $\rvz_\gG^{(0)}$, to highlight the timestep information of the latent diffusion model:
\begin{equation}
\begin{aligned}
    & \bb{E}_{\qphi(\rvz_\gG^{(0)} | \gG)}[\log p_\theta(\rvz_\gG^{(0)})]\\
    = & \bb{E}_{\qphi(\rvz_\gG^{(0)} | \gG)}[\log \int_{\rvz_\gG^{(1:T)}} \big( p_\theta(\rvz_\gG^{({T})}) \prod_{t=1}^{T} p_\theta(\rvz_\gG^{({t-1})} | \rvz_\gG^{(t)}) \big)] \\
    \geq \ & \bb{E}_{\rvz_\gG^{({0:T})}}[  \log p_\theta(\rvz_\gG^{({T})}) + \sum_{t=1}^{T} \log p_\theta(\rvz_\gG^{({t-1})} | \rvz_\gG^{(t)}) - \log q(\rvz_\gG^{({1:T})}|\rvz_\gG^{(0)}) ] \\
    \geq \ & \bb{E}_{\rvz_\gG^{({0:T})}}\Big[  \log p_\theta(\rvz_\gG^{({T})}) -  \log q(\rvz_\gG^{({T})} | \rvz_\gG^{({0})}) - \sum_{t=2}^{T} \underbrace{\KL(q(\rvz_\gG^{({t-1})} | \rvz_\gG^{(t)}, \rvz_\gG^{(0)}) \Vert  p_\theta(\rvz_\gG^{({t-1})} | \rvz_\gG^{(t)}))}_{\gL_\textit{LDM}^{(t-1)}} + \log p_\theta(\rvz_\gG^{({0})} | \rvz_\gG^{(1)})  \Big],
\end{aligned}
\end{equation}
where we factorize $\log p_\theta$ into a sequence of KL divergences between $q(\rvz_\gG^{({t-1})} | \rvz_\gG^{(t)}, \rvz_\gG^{(0)})$ and $p_\theta(\rvz_\gG^{({t-1})} | \rvz_\gG^{(t)})$. Now, for $t \geq 2$, let us consider transitions $q$ and $\ptheta$ with the form in \cref{eq:ddpm_diffusion,eq:ddpm_denoising} respectively, which are both Gaussian distributions with fixed variances. Then we can just set the standard deviation of $p_\theta(\rvx^{({t-1})} | \rvx^{({t})})$ to be the same as that of $q(\rvx^{({t-1})} | \rvx^{({t})}, \rvx^{({0})}))$. With this parameterization, the KL divergence for $\gL_\textit{LDM}^{(t-1)}$ is between two Gaussians with the same standard deviations and thus can be simply calculated as a weighted Euclidean distance between the means. Using the derivation results from in \cref{ssec:background:diffusion} that $\mathbf{\mu}_\theta(\rvx_t,t):=\tfrac{1}{\sqrt{1-\beta_t}}\big(\rvx_t-\tfrac{\beta_t}{\sqrt{1-\alpha_t^2}}\vepsilon_\theta(\rvx_t,t)\big)$, we have that:
\begin{align}
    \gL_\textit{LDM}^{(t-1)} = \bb{E}_{\rvz_0, \vepsilon \sim \mathcal{N} (0, \mI)} \left[\frac{\beta_t^2}{2\rho_t^2(1-\beta_t)(1-\alpha_t^2)} \norm{\vepsilon - { \vepsilon_\theta(\rvz_\gG^{({t})}, t)}}_2^2\right]
\end{align}
which gives us the weights of $w(t)$ for $t = 1,\cdots, T$. For $p_\theta(\rvz_\gG^{({0})} | \rvz_\gG^{(1)})$, we can directly analyze it in the Gaussian form with mean
$$
\mu_\theta(\rvz_\gG^{(1)}, 1) = \frac{\rvz_\gG^{(1)} - \sigma_1 \vepsilon_\theta(\rvz_\gG^{({1})}, 1)}{{\alpha_1}}.
$$
And with $$\rvz_\gG^{(0)} = \frac{\rvz_\gG^{(1)} - \sigma_0 \vepsilon}{{\alpha_1}}$$ we have that:
\begin{align}
    \log p_\theta(\rvz_\gG^{({0})} | \rvz_\gG^{(1)}) = - \log Z^{-1} + \norm{\vepsilon - { \vepsilon_\theta(\rvz_\gG^{({1})},1)}}_2^2 = - \gL_\textit{LDM}^{(0)}
\end{align}
with the normalization constant Z. This distribution gives us the weight of $w(0)$. Besides, we have:
\begin{align}
    \bb{E}_{\rvz_\gG^{({0:T})}}[  \log p_\theta(\rvz_\gG^{({T})}) - q(\rvz_\gG^{({T})} | \rvz_\gG^{({0})})] = 0
\end{align}
since $\rvz_\gG^{({T})} \sim \gN(0, \mI)$ for both $\ptheta$ and $q$. Therefore, without the constants, we have that:
\begin{equation}
\begin{aligned}
    \bb{E}_{\qphi(\rvz_\gG^{(0)} | \gG)}[\log p_\theta(\rvz_\gG^{(0)})] & = \sum_{t=2}^{T} \underbrace{\KL(q(\rvz_\gG^{({t-1})} | \rvz_\gG^{(t)}, \rvz_\gG^{(0)}) \Vert  p_\theta(\rvz_\gG^{({t-1})} | \rvz_\gG^{(t)}))}_{\gL_\textit{LDM}^{(t-1)}} - \log p_\theta(\rvz_\gG^{({0})} | \rvz_\gG^{(1)})\\
    & \geq - \sum_{t=2}^T \gL_\textit{LDM}^{(t-1)} - \gL_\textit{LDM}^{(0)} = -\gL_\textit{LDM}
\end{aligned}
\end{equation}
which completes our proof.
\end{proof}
\begin{proof}[Proof of \cref{theorem:elbo} (\cref{app:eq:loss-invaraince})] Here we show that our derived lower bound is an SE(3)-invariant lower bound. Recall the objective function:
\begin{equation}
\begin{aligned}
    \gL(\rvx, \rvh; \theta, \phi, \xi) \mathrel{\mathop:}= &\underbrace{\bb{E}_{\pdata(\gG), \qphi(\rvz_{\rx}, \rvz_{\rh} | \rvx, \rvh)}\big[\log p_\xi(\rvx, \rvh | \rvz_{\rx}, \rvz_{\rh})\big]}_{\gL_\textit{recon}(\rvx, \rvh; \phi, \xi)} + \\
    &\underbrace{\sum_{t=2}^{T} \KL(q(\rvz_\gG^{({t-1})} | \rvz_\gG^{(t)}, \rvz_\gG^{(0)}) \Vert  p_\theta(\rvz_\gG^{({t-1})} | \rvz_\gG^{(t)})) - \log p_\theta(\rvz_\gG^{({0})} | \rvz_\gG^{(1)})
    }_{\gL_\textit{LDM}(\rvz_{\rx}, \rvz_{\rh}; \theta)}.
\end{aligned}
\end{equation}
Note that, we have $\qphi(\rvz_{\rx}, \rvz_{\rh} | \gG)$ and $\log p_\xi(\rvx, \rvh | \rvz_{\rx}, \rvz_{\rh})$ are equivariant distributions, \textit{i.e.}, $\qphi(\rmR\rvz_{\rx}, \rvz_{\rh} | \rmR \rvx, \rvh)$ and $\log p_\xi(\rmR\rvx, \rvh | \rmR\rvz_{\rx}, \rvz_{\rh})$  for all orthogonal $\rmR$. Then for $\gL_\textit{recon}(\rvx, \rvh)$, we have:
\begin{equation}
\begin{aligned}
    \gL_\textit{recon}(\rmR \rvx, \rvh) & = \bb{E}_{\pdata(\gG), \qphi(\rvz_{\rx}, \rvz_{\rh} | \rmR \rvx, \rvh)}\big[\log p_\xi(\rmR \rvx, \rvh | \rvz_{\rx}, \rvz_{\rh})\big] \\
    & = \int_{\gG} \qphi(\rvz_{\rx}, \rvz_{\rh} | \rmR \rvx, \rvh) \log p_\xi(\rmR \rvx, \rvh | \rvz_{\rx}, \rvz_{\rh}) \\
    & = \int_{\gG} \qphi( \rmR \rmR^{-1} \rvz_{\rx}, \rvz_{\rh} | \rmR \rvx, \rvh) \log p_\xi(\rmR \rvx, \rvh | \rmR \rmR^{-1} \rvz_{\rx}, \rvz_{\rh})  &\text{Multiply by $\rmR \rmR^{-1} = \rmI$} \\
    & = \int_{\gG} \qphi( \rmR^{-1} \rvz_{\rx}, \rvz_{\rh} | \rvx, \rvh) \log p_\xi( \rvx, \rvh | \rmR^{-1} \rvz_{\rx}, \rvz_{\rh})  &\text{Equivariance \& Invariance} \\
    & = \int_{\gG} \qphi( \rvy, \rvz_{\rh} | \rvx, \rvh) \log p_\xi( \rvx, \rvh | \rvy, \rvz_{\rh}) \cdot \underbrace{\det \rmR}_{=1} &\text{Change of Variables $\rvy = \rmR^{-1} \rvz_\rx$} \\
    & = \bb{E}_{\pdata(\gG), \qphi(\rvy, \rvz_{\rh} | \rvx, \rvh)}\big[\log p_\xi(\rvx, \rvh | \rvy, \rvz_{\rh})\big] \\
    & = \gL_\textit{recon}(\rvx, \rvh),
\end{aligned}
\end{equation}
which shows that $\gL_\textit{recon}(\rvx, \rvh)$ is invariant. And for $\gL_\textit{LDM}(\rvz_{\rx}, \rvz_{\rh})$, given that $q(\rvz_\gG^{({t-1})} | \rvz_\gG^{(t)}, \rvz_\gG^{(0)})$ and $p_\theta(\rvz_\gG^{({t-1})} | \rvz_\gG^{(t)})$ are equivariant distributions, we have that:
\begin{align}
    \gL_\textit{LDM}(\rmR \rvz_{\rx}^{(0)}, \rvz_{\rh}^{(0)}) = & \bb{E}_{\pdata(\gG)} \Big[ \sum_{t=2}^{T} \KL(q(\rvz_\rvx^{({t-1})},\rvz_\rvh^{({t-1})} | \rvz_\rvx^{(t)},\rvz_\rvh^{({t})}, \rmR \rvz_\rvx^{(0)},\rvz_\rvh^{({0})}) \Vert  p_\theta(\rvz_\rvx^{({t-1})},\rvz_\rvh^{({t-1})} | \rvz_\rvx^{(t)},\rvz_\rvh^{({t})})) \nonumber \\
    & - \log p_\theta(\rmR \rvz_\rvx^{({0})},\rvz_\rvh^{({0})} | \rvz_\rvx^{(1)},\rvz_\rvh^{({1})}) \Big] \nonumber \\
    = & \int_{\rvz_\gG} \Big[ \sum_{t=2}^{T} \log \frac{q(\rvz_\rvx^{({t-1})},\rvz_\rvh^{({t-1})} | \rvz_\rvx^{(t)},\rvz_\rvh^{({t})}, \rmR \rvz_\rvx^{(0)},\rvz_\rvh^{({0})})}{p_\theta(\rvz_\rvx^{({t-1})} | \rvz_\rvx^{(t)},\rvz_\rvh^{({t})}))} - \log p_\theta(\rmR \rvz_\rvx^{({0})},\rvz_\rvh^{({0})} | \rvz_\rvx^{(1)},\rvz_\rvh^{({1})}) \Big]\nonumber\\
    = & \int_{\rvz_\gG} \Big[ \sum_{t=2}^{T} \log \frac{q(\rmR\rmR^{-1} \rvz_\rvx^{({t-1})},\rvz_\rvh^{({t-1})} | \rmR\rmR^{-1} \rvz_\rvx^{(t)},\rvz_\rvh^{({t})}, \rmR \rvz_\rvx^{(0)},\rvz_\rvh^{({0})})}{p_\theta(\rmR\rmR^{-1} \rvz_\rvx^{({t-1})},\rvz_\rvh^{({t-1})} | \rmR\rmR^{-1} \rvz_\rvx^{(t)},\rvz_\rvh^{({t})}))} \nonumber \\
    & - \log p_\theta(\rmR \rvz_\rvx^{({0})},\rvz_\rvh^{({0})} | \rvz_\rvx^{(1)},\rmR \rmR^{-1} \rvz_\rvh^{({1})}) \Big] \quad \text{(Multiply by $\rmR \rmR^{-1} = \rmI$)} \\
    = & \int_{\rvz_\gG} \Big[ \sum_{t=2}^{T} \log \frac{q(\rmR^{-1} \rvz_\rvx^{({t-1})},\rvz_\rvh^{({t-1})} | \rmR^{-1} \rvz_\rvx^{(t)},\rvz_\rvh^{({t})},  \rvz_\rvx^{(0)},\rvz_\rvh^{({0})})}{p_\theta(\rmR^{-1} \rvz_\rvx^{({t-1})},\rvz_\rvh^{({t-1})} | \rmR^{-1} \rvz_\rvx^{(t)},\rvz_\rvh^{({t})}))} \nonumber\\
    &- \log p_\theta(\rvz_\rvx^{({0})},\rvz_\rvh^{({0})} | \rmR^{-1} \rvz_\rvx^{(1)},\rvz_\rvh^{({1})}) \Big] \quad \text{(Equivariance \& Invariance)} \nonumber\\
    = & \bb{E}_{\pdata(\gG)} \Big[ \sum_{t=2}^{T} \KL(q( \rvy_\rvx^{({t-1})},\rvz_\rvh^{({t-1})} |  \rvy_\rvx^{(t)},\rvz_\rvh^{({t})},  \rvz_\rvx^{(0)},\rvz_\rvh^{({0})}) \Vert  p_\theta( \rvy_\rvx^{({t-1})},\rvz_\rvh^{({t-1})} |  \rvy_\rvx^{(t)},\rvz_\rvh^{({t})})) \nonumber \\
    & - \log p_\theta( \rvz_\rvx^{({0})},\rvz_\rvh^{({0})} |  \rvy_\rvx^{(1)},\rvz_\rvh^{({1})}) \Big] \nonumber\\
    =& \gL_\textit{LDM}(\rvz_{\rx}^{(0)}, \rvz_{\rh}^{(0)}),\nonumber
\end{align}
which shows that $\gL_\textit{LDM}(\rmR \rvz_{\rx}^{(0)}, \rvz_{\rh}^{(0)})$ is invariant. Furthermore, since we operate on the zero-mean subspace, the objectives are naturally also translationally invariant. Thus, we finish the proof.
\end{proof}

\subsection{Invariant Marginal Distribution: \cref{prop:inv-likelihood}}

We also include proofs for key properties of the equivariant probabilistic diffusion model here to be self-contained~\citep{xu2022geodiff,hoogeboom2022equivariant}. Note that, since here we are interested in the equivariant properties, we omit the trivial scalar inputs $\rvh$ and focus on analyzing tensor features $\rvz$. The proof shows that when the the initial distribution $p(\rvz^{(T)})$ is invariant and transition distributions $p(\rvz_\rx^{(t-1)} | \rvz_\rx^{(t)})$ are equivariant, then the marginal distributions $p(\rvz_\rx^{(t)})$ will be invariant, importantly including $p(\rvz_\rx^{(0)})$. Similarly, with decoder $p(\rvx|\rvz_\rx^{(0)})$ also being equivariant, we can further have that our induced distribution $p(\rvx)$ is invariant. 

\begin{proof}
The justification formally can be derived as follow:

\textbf{Condition}: We are given that $p(\rvz_\rx^T) = \mathcal{N}(\mathbf{0}, \rmI)$ is invariant with respect to rotations, \textit{i.e.}, $p(\rvz_\rx^T) = p(\rmR \rvz_\rx^T)$.

\textbf{Derivation}: For $t \in \{1, \cdots, T\}$, let $p(\rvz_\rx^{t-1} | \rvz_\rx^t)$ be equivariant distribution, \textit{i.e.}, $p(\rvz_\rx^{t-1} | \rvz_\rx^t) = p(\rmR \rvz_\rx^{t-1} | \rmR \rvz_\rx^t)$ for all orthogonal $\rmR$. Assume $p(\rvz_\rx^t)$ to be invariant distribution, \textit{i.e.}, $p(\rvz_\rx^t) = p(\rmR \rvz_\rx^t)$ for all orthogonal $\rmR$, then we have:
\begin{align*}
    p(\rmR \rvz_\rx^{t-1}) &= \int_{\rvz_\rx^t} p(\rmR \rvz_\rx^{t-1} | \rvz_\rx^t) p(\rvz_\rx^t)  &\text{Chain Rule} \\ 
    &= \int_{\rvz_\rx^t} p(\rmR \rvz_\rx^{t-1} | \rmR \rmR^{-1} \rvz_\rx^t) p(\rmR \rmR^{-1} \rvz_\rx^t)  &\text{Multiply by $\rmR \rmR^{-1} = \rmI$} \\ 
    &= \int_{\rvz_\rx^t} p( \rvz_\rx^{t-1} | \rmR^{-1} \rvz_\rx^t) p( \rmR^{-1} \rvz_\rx^t) &\text{Equivariance \& Invariance} \\ 
    &= \int_{\rvy} p( \rvz_\rx^{t-1} | \rvy) p( \rvy) \cdot \underbrace{\det \rmR}_{=1} &\text{Change of Variables $\rvy = \rmR^{-1} \rvz_\rx^t$} \\ 
    &= p(\rvz_\rx^{t-1}),
\end{align*}
and therefore $p(\rvz_\rx^{t-1})$ is invariant. By induction, $p(\rvz_\rx^{T-1}), \ldots, p(\rvz_\rx^{0})$ are all invariant. Furthermore, since the decoder $p(\rvx|\rvz_\rx^{(0)})$ is also equivariant, with the same derivation we also have that our induced distribution $p(\rvx)$ is invariant.
\end{proof}

\section{Model Architecture Details}
\label{app:sec:models}

In our implementation, all models are parameterized with EGNNs~\citep{satorras2021en} as backbone. EGNNs are a class of Graph Neural Network that satisfies the equivariance property in \cref{eq:equivariance:ae,eq:equivariance:ldm}. 
In this work, we consider molecular geometries as point clouds, without specifying the connecting bonds. Therefore, in practice, we take the point clouds as fully connected graph $G$ and model the interactions between all atoms $\rv_i \in \mathcal{V}$.
Each node $\rv_i$ is embedded with coordinates $\rx_i \in \mathbb R^3$ and atomic features $\rh_i \in \mathbb R^d$. 
Then, EGNNs are composed of multiple Equivariant Convolutional Layers $\rvx^{l+1}, \rvh^{l+1} = \operatorname{EGCL}[\rvx^l, \rvh^l]$, with each single layer defined as:
\begin{equation}
\begin{aligned}
    & \rvm_{ij} = \phi_{e}\left(\rh_{i}^{l}, \rh_{j}^{l}, d_{ij}^2, a_{ij}\right), \\
    & \rh_{i}^{l+1} = \phi_{h}(\rh_{i}^l, { \sum_{j \neq i}} \tilde{e}_{ij} \rvm_{ij}), \\
    & \rx_{i}^{l+1} = \rx_{i}^{l}+\sum_{j \neq i} \frac{\rx_{i}^{l}-\rx_{j}^{l}}{d_{ij}+ 1} \phi_{x}\left(\rh_{i}^{l}, \rh_{j}^{l}, d_{ij}^2, a_{ij}\right),
\label{eq:egnn-layer} 
\end{aligned}
\end{equation}
where $l$ denotes the layer index. $\tilde{e}_{ij}=\phi_{inf}(\rvm_{ij})$ acts as the attention weights to reweight messages passed from different edges. $d_{ij} = \| \rx_{i}^{l}-\rx_{j}^{l} \|_2$ represents the pairwise distance between atoms $\rv_i$ and $\rv_j$, and $a_{ij}$ are optional edge features. 
We follow previous work~\citep{satorras2021enflow,hoogeboom2022equivariant} to normalize the relative directions $\rx_i^l - \rx_j^l$ in \cref{eq:egnn-layer} by $d_{ij} + 1$, which empirically improved model stability. All learnable functions, \textit{i.e.}, $\phi_e$, $\phi_h$, $\phi_x$ and $\phi_{inf}$, are parameterized by Multi Layer Perceptrons (MLPs). 
Then a complete EGNN model can be realized by stacking $L$ $\operatorname{EGCL}$ layers such that $\rvx^L, \rvh^L = \operatorname{EGNN}[\rvx^0, \rvh^0]$, which can satisfy the required equivariant constraint in \cref{eq:equivariance:ae,eq:equivariance:ldm}.

\section{Featurization and Implementation Details}
\label{app:sec:exp-details}

We use the open-source software \textsc{RDkit}~\citep{landrum2016rdkit} to preprocess molecules.
For QM9 we take atom types (H, C, N, O, F) and integer-valued atom charges as atomic features, while for Drugs we only use atom types. 
The results reported in \cref{ssec:exp-molgen,ssec:exp-condition} are based on the \textit{ES-reg} regularization strategy (\cref{ssec:method:train}), where the encoder is only optimized with $1000$ iterations of warm-up training and then fixed.
For the diffusion process~(\cref{eq:ddpm_diffusion}), we use the polynomial noise schedule~\citep{hoogeboom2022equivariant,wu2022diffusionbased}, where $\alpha$ linearly decays from $10^3/T$ to $0$ \textit{w.r.t.} time step $t$. 
And for the denoising process~(\cref{eq:ddpm_denoising}), the variances are defined as $\rho_t = \sqrt{\frac{\sigma_{t-1}}{\sigma_t}}\beta_t$.

All neural networks used for the encoder, latent diffusion, and decoder are implemented with EGNNs~\cite{satorras2021en} by PyTorch~\citep{pytorch2017automatic} package, as introduced in \cref{app:sec:models}. We set the dimension of latent invariant features $k$ to $1$ for QM9 and $2$ for DRUG, which extremely reduces the atomic feature dimension. 
For the training of latent denoising network $\vepsilon_\theta$: on QM9, we train EGNNs with $9$ layers and $256$ hidden features with a batch size $64$; and on GEOM-DRUG, we train EGNNs with $4$ layers and $256$ hidden features, with batch size $64$. 
For the autoencoders, we parameterize the decoder $\gD_\xi$ in the same way as $\vepsilon_\theta$, but implement the encoder $\gE_\phi$ with a $1$ layer EGNN. The shallow encoder in practice constrains the encoding capacity and helps regularize the latent space. All models use SiLU activations. 
We train all the modules until convergence.
For all the experiments, we choose the Adam optimizer ~\citep{kingma2014adam} with a constant learning rate of $10^{-4}$ as our default training configuration.
The training on QM9 takes approximately $2000$ epochs, and on DRUG takes $20$ epochs. 

\section{Ablation Studies}
\label{app:sec:exp-ablation}

In this section, We provide additional experimental results on QM9 to justify the effect of several model designs. Specifically, we perform ablation studies on two key model designs: \textit{autoencoder regularization method} and \textit{latent space dimension $k$}. The results are reported in \cref{app:tab:ablation}.

\begin{table*}[!ht]
\centering
\caption{Results of ablation study with different model designs. Metrics are calculated with 10000 samples generated from each setting.}
\label{app:tab:ablation}
\begin{threeparttable}
\begin{tabular}{l | c c c c }
    \toprule[1.0pt]
    \# Metrics & Atom Sta (\%) & Mol Sta (\%) & Valid (\%) & Valid \& Unique (\%) \\
    \midrule[0.8pt]
    \method ($k=1$, \textit{KL-reg})* &  95.45 & 40.7 & 83.7 & 83.5 \\
    \method ($k=16$, \textit{ES-reg}) & 98.6 & 86.0 & 92.4 & 92.2 \\
    \method ($k=8$, \textit{ES-reg}) & 98.7 & 87.1 & 92.1 & 92.0 \\
    \method ($k=4$, \textit{ES-reg}) & 98.8 & 87.4 & 92.6 & 92.5 \\ 
    \rowcolor{lightgray} \method ($k=1$, \textit{ES-reg}) & \textbf{98.9} $\pm$ 0.1 & \textbf{89.4} $\pm$ 0.5 & \textbf{93.8} $\pm$ 0.4 & \textbf{92.7} $\pm$ 0.5  \\
    \bottomrule[1.0pt]
\end{tabular}
\begin{tablenotes}
\small
\item *Note that this reported result is already the best result we achieved for \textit{KL-reg}.
\end{tablenotes}
\end{threeparttable}
\end{table*}

We first discuss the effect of different autoencoder regularization methods, \textit{i.e.}, \textit{KL-reg} and \textit{ES-reg} (see details in \ref{ssec:method:autoencoder}), with the latent invariant feature dimension fixed as $1$. 
Following previous practice in latent diffusion models of image and point clouds domains~\cite{rombach2022high,zeng2022lion}, for \textit{KL-reg}, we also weight the KL term with a small factor $0.01$. However, during our initial experiments where we naturally first try the \textit{KL-reg} method, we observed unexpected failure with extremely poor performance, as shown in the first row in \cref{app:tab:ablation}. Note that, this reported result is already the best result we achieved for \textit{KL-reg}, with searching over a large range of KL term weights and latent space dimension. In practice, we even notice the \textit{KL-reg} is unstable for training, which often suffers from numerical errors during training. Our closer observation of the experimental results suggests that the \textit{equivariant latent feature} part always tends to converge to highly scattered means and extremely small variances, which leads to the numerical issue for calculating KL term and also is not suitable for LDM training. Therefore, we turned to constraining the encoder, more precisely, constraining the value scale of encoded latent features, by early stopping the training encoder. This easy strategy turned out to work pretty well in practice as shown in \cref{app:tab:ablation}, and we leave the further study of \textit{KL-reg} as future work in this area.

We further study the effect of latent invariance feature dimension $k$, and the results are also reported in \cref{app:tab:ablation}. As shown in the table, we observe that generally \method shows better performance with lower $k$. This phenomenon verifies our motivation that a lower dimensionality can alleviate the generative modeling complexity and benefit the training of LDM. Specifically, the performances of \method on QM9 with $k$ set as $1$ or $2$ are very similar, so we only report $k=1$ as representative in \cref{app:tab:ablation}. In practice, we set $k$ as $1$ for QM9 dataset and $2$ for DRUG which contains more atom types.

\section{More Visualization Results}
\label{app:sec:vis}

\begin{figure}[!ht]
    \centering
    \includegraphics[width=1.0\linewidth]{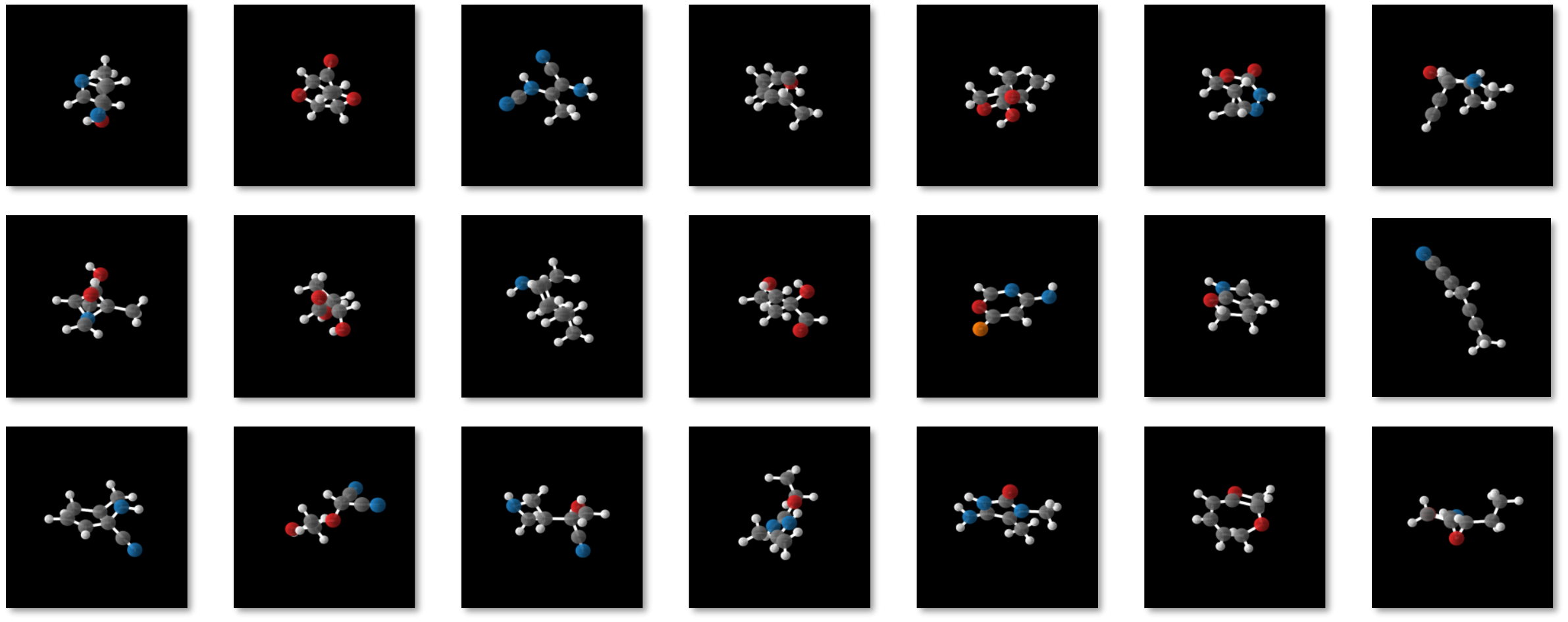}
    \vspace{-15pt}
    \caption{Molecules generated from \method trained on QM9.}
    \label{app:fig:vis-qm9}
\end{figure}

\begin{figure}[!ht]
    \centering
    \includegraphics[width=1.0\linewidth]{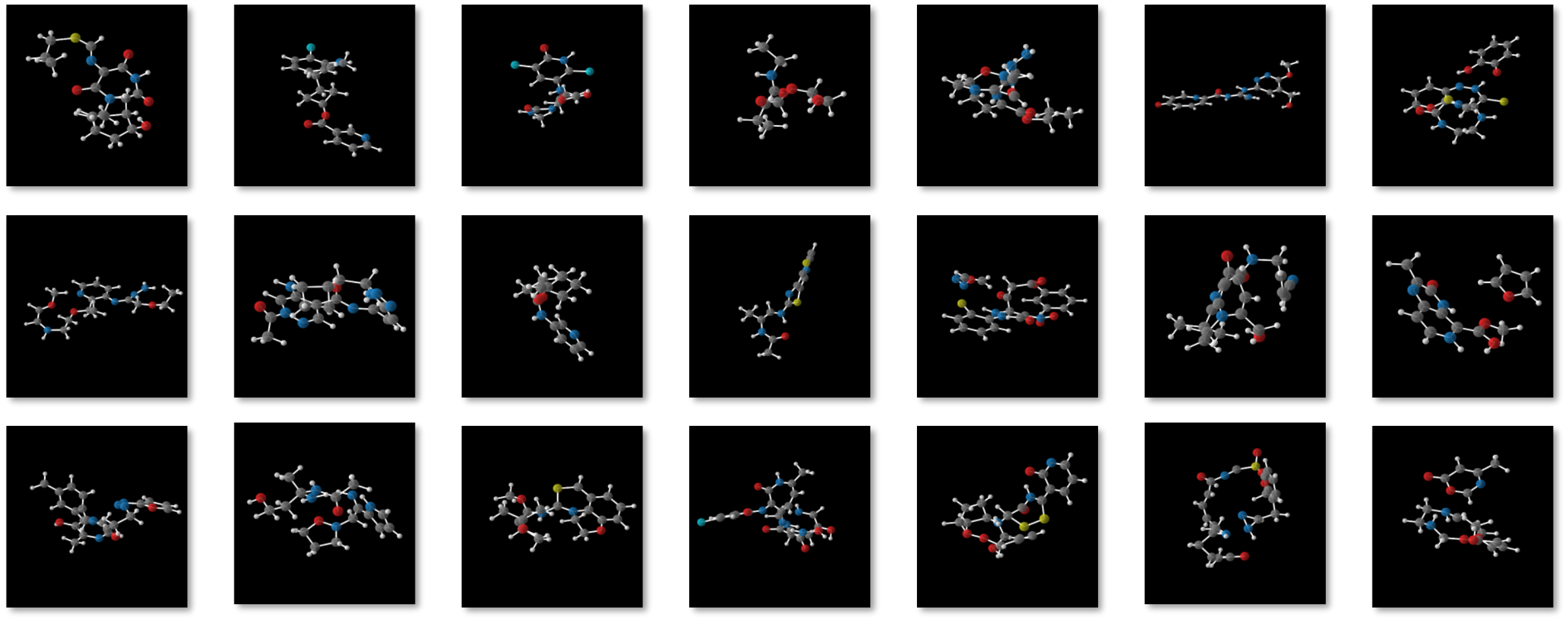}
    \vspace{-15pt}
    \caption{Molecules generated from \method trained on DRUG.}
    \label{app:fig:vis-drug}
\end{figure}

In this section, we provide more visualizations of molecules generated from \method. Samples drawn from models trained on QM9 and DRUG are provided in \cref{app:fig:vis-qm9} and \cref{app:fig:vis-drug} respectively. These examples are randomly generated without any cherry pick. Therefore, the generated geometries might be difficult to see in some figures due to imperfect viewing direction.

As shown in the two figures, the model is always able to generate realistic molecular geometries for both small and large size molecules. An outlier case is that the model occasionally generates disconnected components, as shown in the rightest column of \cref{app:fig:vis-drug}, which happens more often when trained on the large molecule DRUG dataset. However, this phenomenon actually is not a problem and is common in all non-autoregressive molecule generative models~\cite{zang2020moflow,jo2022score}, and can be easily fixed by just filtering the smaller components.

\end{document}

%% file: math_commands.tex
\usepackage{amsmath,amsfonts,bm,mathtools}

\def\eqref#1{equation~\ref{#1}}

\def\1{\bm{1}}

\def\rh{{\textnormal{h}}}

\def\rv{{\textnormal{v}}}

\def\rx{{\textnormal{x}}}

\def\rvh{{\mathbf{h}}}

\def\rvm{{\mathbf{m}}}

\def\rvx{{\mathbf{x}}}
\def\rvy{{\mathbf{y}}}
\def\rvz{{\mathbf{z}}}

\def\rmI{{\mathbf{I}}}

\def\rmR{{\mathbf{R}}}

\def\rmU{{\mathbf{U}}}

\def\vmu{{\bm{\mu}}}

\def\vepsilon{{\bm{\epsilon}}}

\def\vd{{\bm{d}}}

\def\vt{{\bm{t}}}

\def\mI{{\bm{I}}}

\DeclareMathAlphabet{\mathsfit}{\encodingdefault}{\sfdefault}{m}{sl}
\SetMathAlphabet{\mathsfit}{bold}{\encodingdefault}{\sfdefault}{bx}{n}

\def\gD{{\mathcal{D}}}
\def\gE{{\mathcal{E}}}
\def\gF{{\mathcal{F}}}
\def\gG{{\mathcal{G}}}

\def\gL{{\mathcal{L}}}

\def\gN{{\mathcal{N}}}

\newcommand{\pdata}{p_{\rm{data}}}

\newcommand{\R}{\mathbb{R}}

\newcommand{\KL}{D_{\mathrm{KL}}}

%% file: icml2023.bbl
\begin{thebibliography}{60}
\providecommand{\natexlab}[1]{#1}
\providecommand{\url}[1]{\texttt{#1}}
\expandafter\ifx\csname urlstyle\endcsname\relax
  \providecommand{\doi}[1]{doi: #1}\else
  \providecommand{\doi}{doi: \begingroup \urlstyle{rm}\Url}\fi

\bibitem[Anand \& Achim(2022)Anand and Achim]{anand2022protein}
Anand, N. and Achim, T.
\newblock Protein structure and sequence generation with equivariant denoising
  diffusion probabilistic models.
\newblock \emph{arXiv preprint arXiv:2205.15019}, 2022.

\bibitem[Anderson et~al.(2019)Anderson, Hy, and Kondor]{anderson2019cormorant}
Anderson, B., Hy, T.~S., and Kondor, R.
\newblock Cormorant: Covariant molecular neural networks.
\newblock \emph{Advances in neural information processing systems}, 32, 2019.

\bibitem[Aneja et~al.(2021)Aneja, Schwing, Kautz, and
  Vahdat]{aneja2021contrastive}
Aneja, J., Schwing, A., Kautz, J., and Vahdat, A.
\newblock A contrastive learning approach for training variational autoencoder
  priors.
\newblock \emph{Advances in neural information processing systems},
  34:\penalty0 480--493, 2021.

\bibitem[Batzner et~al.(2021)Batzner, Smidt, Sun, Mailoa, Kornbluth, Molinari,
  and Kozinsky]{batzner2021se}
Batzner, S., Smidt, T.~E., Sun, L., Mailoa, J.~P., Kornbluth, M., Molinari, N.,
  and Kozinsky, B.
\newblock Se (3)-equivariant graph neural networks for data-efficient and
  accurate interatomic potentials.
\newblock \emph{arXiv preprint arXiv:2101.03164}, 2021.

\bibitem[Dai \& Wipf(2019)Dai and Wipf]{dai2018diagnosing}
Dai, B. and Wipf, D.
\newblock Diagnosing and enhancing {VAE} models.
\newblock In \emph{International Conference on Learning Representations}, 2019.
\newblock URL \url{https://openreview.net/forum?id=B1e0X3C9tQ}.

\bibitem[Esser et~al.(2021)Esser, Rombach, and Ommer]{esser2021taming}
Esser, P., Rombach, R., and Ommer, B.
\newblock Taming transformers for high-resolution image synthesis.
\newblock In \emph{Proceedings of the IEEE/CVF conference on computer vision
  and pattern recognition}, pp.\  12873--12883, 2021.

\bibitem[Fuchs et~al.(2020)Fuchs, Worrall, Fischer, and Welling]{fuchs2020se3}
Fuchs, F., Worrall, D., Fischer, V., and Welling, M.
\newblock Se(3)-transformers: 3d roto-translation equivariant attention
  networks.
\newblock \emph{NeurIPS}, 2020.

\bibitem[Gebauer et~al.(2019)Gebauer, Gastegger, and
  Sch{\"u}tt]{gebauer2019symmetry}
Gebauer, N., Gastegger, M., and Sch{\"u}tt, K.
\newblock Symmetry-adapted generation of 3d point sets for the targeted
  discovery of molecules.
\newblock \emph{Advances in neural information processing systems}, 32, 2019.

\bibitem[Gebauer et~al.(2021)Gebauer, Gastegger, Hessmann, M{\"u}ller, and
  Sch{\"u}tt]{gebauer2021inverse}
Gebauer, N.~W., Gastegger, M., Hessmann, S.~S., M{\"u}ller, K.-R., and
  Sch{\"u}tt, K.~T.
\newblock Inverse design of 3d molecular structures with conditional generative
  neural networks.
\newblock \emph{arXiv preprint arXiv:2109.04824}, 2021.

\bibitem[Ghosh et~al.(2020)Ghosh, Sajjadi, Vergari, Black, and
  Scholkopf]{Ghosh2020From}
Ghosh, P., Sajjadi, M. S.~M., Vergari, A., Black, M., and Scholkopf, B.
\newblock From variational to deterministic autoencoders.
\newblock In \emph{International Conference on Learning Representations}, 2020.
\newblock URL \url{https://openreview.net/forum?id=S1g7tpEYDS}.

\bibitem[Graves et~al.(2020)Graves, Byerly, Priego, Makkapati, Parish,
  Medellin, and Berrondo]{graves2020review}
Graves, J., Byerly, J., Priego, E., Makkapati, N., Parish, S.~V., Medellin, B.,
  and Berrondo, M.
\newblock A review of deep learning methods for antibodies.
\newblock \emph{Antibodies}, 9\penalty0 (2):\penalty0 12, 2020.

\bibitem[Ho et~al.(2020)Ho, Jain, and Abbeel]{ho2020denoising}
Ho, J., Jain, A., and Abbeel, P.
\newblock Denoising diffusion probabilistic models.
\newblock \emph{arXiv preprint arXiv:2006.11239}, 2020.

\bibitem[Hoogeboom et~al.(2022)Hoogeboom, Satorras, Vignac, and
  Welling]{hoogeboom2022equivariant}
Hoogeboom, E., Satorras, V.~G., Vignac, C., and Welling, M.
\newblock Equivariant diffusion for molecule generation in 3d.
\newblock In \emph{International Conference on Machine Learning}, pp.\
  8867--8887. PMLR, 2022.

\bibitem[Jin et~al.(2018)Jin, Barzilay, and Jaakkola]{jin2018junction}
Jin, W., Barzilay, R., and Jaakkola, T.
\newblock Junction tree variational autoencoder for molecular graph generation.
\newblock \emph{arXiv preprint arXiv:1802.04364}, 2018.

\bibitem[Jing et~al.(2021)Jing, Eismann, Suriana, Townshend, and
  Dror]{jing2021gvp}
Jing, B., Eismann, S., Suriana, P., Townshend, R. J.~L., and Dror, R.
\newblock Learning from protein structure with geometric vector perceptrons.
\newblock In \emph{International Conference on Learning Representations}, 2021.

\bibitem[Jo et~al.(2022)Jo, Lee, and Hwang]{jo2022score}
Jo, J., Lee, S., and Hwang, S.~J.
\newblock Score-based generative modeling of graphs via the system of
  stochastic differential equations.
\newblock \emph{arXiv preprint arXiv:2202.02514}, 2022.

\bibitem[Kingma \& Ba(2014)Kingma and Ba]{kingma2014adam}
Kingma, D.~P. and Ba, J.
\newblock Adam: A method for stochastic optimization.
\newblock In \emph{3nd International Conference on Learning Representations},
  2014.

\bibitem[Kingma \& Welling(2013)Kingma and Welling]{kingma2013auto}
Kingma, D.~P. and Welling, M.
\newblock Auto-encoding variational bayes.
\newblock In \emph{2nd International Conference on Learning Representations},
  2013.

\bibitem[Kong et~al.(2021)Kong, Ping, Huang, Zhao, and
  Catanzaro]{kong2021diffwave}
Kong, Z., Ping, W., Huang, J., Zhao, K., and Catanzaro, B.
\newblock Diffwave: A versatile diffusion model for audio synthesis.
\newblock In \emph{International Conference on Learning Representations}, 2021.

\bibitem[Landrum(2016)]{landrum2016rdkit}
Landrum, G.
\newblock Rdkit: open-source cheminformatics http://www. rdkit. org.
\newblock 2016.

\bibitem[Li et~al.(2022)Li, Thickstun, Gulrajani, Liang, and
  Hashimoto]{li2022diffusionlm}
Li, X.~L., Thickstun, J., Gulrajani, I., Liang, P., and Hashimoto, T.
\newblock Diffusion-{LM} improves controllable text generation.
\newblock In Oh, A.~H., Agarwal, A., Belgrave, D., and Cho, K. (eds.),
  \emph{Advances in Neural Information Processing Systems}, 2022.
\newblock URL \url{https://openreview.net/forum?id=3s9IrEsjLyk}.

\bibitem[Li et~al.(2021)Li, Pei, and Lai]{li2021structure}
Li, Y., Pei, J., and Lai, L.
\newblock Structure-based de novo drug design using 3d deep generative models.
\newblock \emph{Chemical science}, 12\penalty0 (41):\penalty0 13664--13675,
  2021.

\bibitem[Lin et~al.(2022)Lin, Huang, Liu, Li, Ji, and Li]{lin2022diffbp}
Lin, H., Huang, Y., Liu, M., Li, X., Ji, S., and Li, S.~Z.
\newblock Diffbp: Generative diffusion of 3d molecules for target protein
  binding.
\newblock \emph{arXiv preprint arXiv:2211.11214}, 2022.

\bibitem[Liu et~al.(2018)Liu, Allamanis, Brockschmidt, and
  Gaunt]{Liu2019constrained}
Liu, Q., Allamanis, M., Brockschmidt, M., and Gaunt, A.
\newblock Constrained graph variational autoencoders for molecule design.
\newblock In \emph{Advances in neural information processing systems}, 2018.

\bibitem[Luo et~al.(2022)Luo, Su, Peng, Wang, Peng, and
  Ma]{luo2022antigenspecific}
Luo, S., Su, Y., Peng, X., Wang, S., Peng, J., and Ma, J.
\newblock Antigen-specific antibody design and optimization with
  diffusion-based generative models for protein structures.
\newblock In Oh, A.~H., Agarwal, A., Belgrave, D., and Cho, K. (eds.),
  \emph{Advances in Neural Information Processing Systems}, 2022.
\newblock URL \url{https://openreview.net/forum?id=jSorGn2Tjg}.

\bibitem[Luo \& Ji(2021)Luo and Ji]{luo2021autoregressive}
Luo, Y. and Ji, S.
\newblock An autoregressive flow model for 3d molecular geometry generation
  from scratch.
\newblock In \emph{International Conference on Learning Representations}, 2021.

\bibitem[Ma et~al.(2019)Ma, Zhou, Li, Neubig, and Hovy]{ma2019flowseq}
Ma, X., Zhou, C., Li, X., Neubig, G., and Hovy, E.
\newblock {F}low{S}eq: Non-autoregressive conditional sequence generation with
  generative flow.
\newblock In \emph{Proceedings of the 2019 Conference on Empirical Methods in
  Natural Language Processing and the 9th International Joint Conference on
  Natural Language Processing (EMNLP-IJCNLP)}, pp.\  4282--4292, Hong Kong,
  China, November 2019. Association for Computational Linguistics.
\newblock \doi{10.18653/v1/D19-1437}.
\newblock URL \url{https://aclanthology.org/D19-1437}.

\bibitem[Masuda et~al.(2020)Masuda, Ragoza, and Koes]{masuda2020generating}
Masuda, T., Ragoza, M., and Koes, D.~R.
\newblock Generating 3d molecular structures conditional on a receptor binding
  site with deep generative models.
\newblock \emph{arXiv preprint arXiv:2010.14442}, 2020.

\bibitem[Meng et~al.(2022)Meng, He, Song, Song, Wu, Zhu, and
  Ermon]{meng2022sdedit}
Meng, C., He, Y., Song, Y., Song, J., Wu, J., Zhu, J.-Y., and Ermon, S.
\newblock {SDE}dit: Guided image synthesis and editing with stochastic
  differential equations.
\newblock In \emph{International Conference on Learning Representations}, 2022.
\newblock URL \url{https://openreview.net/forum?id=aBsCjcPu_tE}.

\bibitem[Nichol \& Dhariwal(2021)Nichol and Dhariwal]{nichol2021improved}
Nichol, A.~Q. and Dhariwal, P.
\newblock Improved denoising diffusion probabilistic models.
\newblock In \emph{International Conference on Machine Learning}, pp.\
  8162--8171. PMLR, 2021.

\bibitem[Paszke et~al.(2017)Paszke, Gross, Chintala, Chanan, Yang, DeVito, Lin,
  Desmaison, Antiga, and Lerer]{pytorch2017automatic}
Paszke, A., Gross, S., Chintala, S., Chanan, G., Yang, E., DeVito, Z., Lin, Z.,
  Desmaison, A., Antiga, L., and Lerer, A.
\newblock Automatic differentiation in pytorch.
\newblock In \emph{NIPS-W}, 2017.

\bibitem[Peng et~al.(2022)Peng, Luo, Guan, Xie, Peng, and Ma]{Pocket2Mol}
Peng, X., Luo, S., Guan, J., Xie, Q., Peng, J., and Ma, J.
\newblock Pocket2mol: Efficient molecular sampling based on 3d protein pockets.
\newblock In \emph{International Conference on Machine Learning}, 2022.

\bibitem[Pereira et~al.(2016)Pereira, Caffarena, and
  Dos~Santos]{pereira2016boosting}
Pereira, J.~C., Caffarena, E.~R., and Dos~Santos, C.~N.
\newblock Boosting docking-based virtual screening with deep learning.
\newblock \emph{Journal of chemical information and modeling}, 56\penalty0
  (12):\penalty0 2495--2506, 2016.

\bibitem[Powers et~al.(2022)Powers, Yu, Suriana, and Dror]{Powers2022}
Powers, A.~S., Yu, H.~H., Suriana, P., and Dror, R.~O.
\newblock Fragment-based ligand generation guided by geometric deep learning on
  protein-ligand structure.
\newblock \emph{bioRxiv}, 2022.
\newblock \doi{10.1101/2022.03.17.484653}.
\newblock URL
  \url{https://www.biorxiv.org/content/early/2022/03/21/2022.03.17.484653}.

\bibitem[Ramakrishnan et~al.(2014)Ramakrishnan, Dral, Rupp, and
  Von~Lilienfeld]{ramakrishnan2014quantum}
Ramakrishnan, R., Dral, P.~O., Rupp, M., and Von~Lilienfeld, O.~A.
\newblock Quantum chemistry structures and properties of 134 kilo molecules.
\newblock \emph{Scientific data}, 1\penalty0 (1):\penalty0 1--7, 2014.

\bibitem[Ramesh et~al.(2022)Ramesh, Dhariwal, Nichol, Chu, and
  Chen]{ramesh2022hierarchical}
Ramesh, A., Dhariwal, P., Nichol, A., Chu, C., and Chen, M.
\newblock Hierarchical text-conditional image generation with clip latents.
\newblock \emph{arXiv preprint arXiv:2204.06125}, 2022.

\bibitem[Razavi et~al.(2019)Razavi, Van~den Oord, and
  Vinyals]{razavi2019generating}
Razavi, A., Van~den Oord, A., and Vinyals, O.
\newblock Generating diverse high-fidelity images with vq-vae-2.
\newblock \emph{Advances in neural information processing systems}, 32, 2019.

\bibitem[Rombach et~al.(2022)Rombach, Blattmann, Lorenz, Esser, and
  Ommer]{rombach2022high}
Rombach, R., Blattmann, A., Lorenz, D., Esser, P., and Ommer, B.
\newblock High-resolution image synthesis with latent diffusion models.
\newblock In \emph{Proceedings of the IEEE/CVF Conference on Computer Vision
  and Pattern Recognition}, pp.\  10684--10695, 2022.

\bibitem[Satorras et~al.(2021{\natexlab{a}})Satorras, Hoogeboom, Fuchs, Posner,
  and Welling]{satorras2021enflow}
Satorras, V.~G., Hoogeboom, E., Fuchs, F.~B., Posner, I., and Welling, M.
\newblock E (n) equivariant normalizing flows for molecule generation in 3d.
\newblock \emph{arXiv preprint arXiv:2105.09016}, 2021{\natexlab{a}}.

\bibitem[Satorras et~al.(2021{\natexlab{b}})Satorras, Hoogeboom, and
  Welling]{satorras2021en}
Satorras, V.~G., Hoogeboom, E., and Welling, M.
\newblock E(n) equivariant graph neural networks.
\newblock In \emph{International conference on machine learning}, pp.\
  9323--9332. PMLR, 2021{\natexlab{b}}.

\bibitem[Sch\"{u}tt et~al.(2017)Sch\"{u}tt, Kindermans, Sauceda~Felix, Chmiela,
  Tkatchenko, and M\"{u}ller]{schutt2017schnet}
Sch\"{u}tt, K., Kindermans, P.-J., Sauceda~Felix, H.~E., Chmiela, S.,
  Tkatchenko, A., and M\"{u}ller, K.-R.
\newblock Schnet: A continuous-filter convolutional neural network for modeling
  quantum interactions.
\newblock In \emph{Advances in Neural Information Processing Systems}, pp.\
  991--1001. Curran Associates, Inc., 2017.

\bibitem[Serre et~al.(1977)]{serre1977linear}
Serre, J.-P. et~al.
\newblock \emph{Linear representations of finite groups}, volume~42.
\newblock Springer, 1977.

\bibitem[Shi et~al.(2020)Shi, Xu, Zhu, Zhang, Zhang, and Tang]{shi2020graphaf}
Shi, C., Xu, M., Zhu, Z., Zhang, W., Zhang, M., and Tang, J.
\newblock Graphaf: a flow-based autoregressive model for molecular graph
  generation.
\newblock \emph{arXiv preprint arXiv:2001.09382}, 2020.

\bibitem[Sinha et~al.(2021)Sinha, Song, Meng, and Ermon]{sinha2021d2c}
Sinha, A., Song, J., Meng, C., and Ermon, S.
\newblock D2c: Diffusion-decoding models for few-shot conditional generation.
\newblock \emph{Advances in Neural Information Processing Systems},
  34:\penalty0 12533--12548, 2021.

\bibitem[Sohl-Dickstein et~al.(2015)Sohl-Dickstein, Weiss, Maheswaranathan, and
  Ganguli]{sohl2015deep}
Sohl-Dickstein, J., Weiss, E.~A., Maheswaranathan, N., and Ganguli, S.
\newblock Deep unsupervised learning using nonequilibrium thermodynamics.
\newblock \emph{arXiv preprint arXiv:1503.03585}, 2015.

\bibitem[Song \& Ermon(2019)Song and Ermon]{song2019generative}
Song, Y. and Ermon, S.
\newblock Generative modeling by estimating gradients of the data distribution.
\newblock In \emph{Advances in Neural Information Processing Systems}, pp.\
  11918--11930, 2019.

\bibitem[Song et~al.(2021)Song, Sohl-Dickstein, Kingma, Kumar, Ermon, and
  Poole]{song2021scorebased}
Song, Y., Sohl-Dickstein, J., Kingma, D.~P., Kumar, A., Ermon, S., and Poole,
  B.
\newblock Score-based generative modeling through stochastic differential
  equations.
\newblock In \emph{International Conference on Learning Representations}, 2021.

\bibitem[Thomas et~al.(2018)Thomas, Smidt, Kearnes, Yang, Li, Kohlhoff, and
  Riley]{Thomas2018TensorFN}
Thomas, N., Smidt, T., Kearnes, S.~M., Yang, L., Li, L., Kohlhoff, K., and
  Riley, P.
\newblock Tensor field networks: Rotation- and translation-equivariant neural
  networks for 3d point clouds.
\newblock \emph{ArXiv}, 2018.

\bibitem[Townshend et~al.(2021)Townshend, V{\"o}gele, Suriana, Derry, Powers,
  Laloudakis, Balachandar, Jing, Anderson, Eismann, Kondor, Altman, and
  Dror]{townshend2021atomd}
Townshend, R. J.~L., V{\"o}gele, M., Suriana, P.~A., Derry, A., Powers, A.,
  Laloudakis, Y., Balachandar, S., Jing, B., Anderson, B.~M., Eismann, S.,
  Kondor, R., Altman, R., and Dror, R.~O.
\newblock {ATOM}3d: Tasks on molecules in three dimensions.
\newblock In \emph{Thirty-fifth Conference on Neural Information Processing
  Systems Datasets and Benchmarks Track (Round 1)}, 2021.
\newblock URL \url{https://openreview.net/forum?id=FkDZLpK1Ml2}.

\bibitem[Trippe et~al.(2022)Trippe, Yim, Tischer, Broderick, Baker, Barzilay,
  and Jaakkola]{trippe2022diffusion}
Trippe, B.~L., Yim, J., Tischer, D., Broderick, T., Baker, D., Barzilay, R.,
  and Jaakkola, T.
\newblock Diffusion probabilistic modeling of protein backbones in 3d for the
  motif-scaffolding problem.
\newblock \emph{arXiv preprint arXiv:2206.04119}, 2022.

\bibitem[Vahdat et~al.(2021)Vahdat, Kreis, and Kautz]{vahdat2021score}
Vahdat, A., Kreis, K., and Kautz, J.
\newblock Score-based generative modeling in latent space.
\newblock \emph{Advances in Neural Information Processing Systems},
  34:\penalty0 11287--11302, 2021.

\bibitem[Van Den~Oord et~al.(2016)Van Den~Oord, Kalchbrenner, and
  Kavukcuoglu]{van2016pixel}
Van Den~Oord, A., Kalchbrenner, N., and Kavukcuoglu, K.
\newblock Pixel recurrent neural networks.
\newblock In \emph{International conference on machine learning}, pp.\
  1747--1756. PMLR, 2016.

\bibitem[Weiler et~al.(2018)Weiler, Geiger, Welling, Boomsma, and
  Cohen]{Weiler20183DSC}
Weiler, M., Geiger, M., Welling, M., Boomsma, W., and Cohen, T.
\newblock 3d steerable cnns: Learning rotationally equivariant features in
  volumetric data.
\newblock In \emph{NeurIPS}, 2018.

\bibitem[Winter et~al.(2021)Winter, No{\'e}, and Clevert]{winter2021auto}
Winter, R., No{\'e}, F., and Clevert, D.-A.
\newblock Auto-encoding molecular conformations.
\newblock \emph{arXiv preprint arXiv:2101.01618}, 2021.

\bibitem[Winter et~al.(2022)Winter, Bertolini, Le, Noe, and
  Clevert]{winter2022unsupervised}
Winter, R., Bertolini, M., Le, T., Noe, F., and Clevert, D.-A.
\newblock Unsupervised learning of group invariant and equivariant
  representations.
\newblock In Oh, A.~H., Agarwal, A., Belgrave, D., and Cho, K. (eds.),
  \emph{Advances in Neural Information Processing Systems}, 2022.
\newblock URL \url{https://openreview.net/forum?id=47lpv23LDPr}.

\bibitem[Wu et~al.(2022)Wu, Gong, Liu, Ye, and qiang liu]{wu2022diffusionbased}
Wu, L., Gong, C., Liu, X., Ye, M., and qiang liu.
\newblock Diffusion-based molecule generation with informative prior bridges.
\newblock In Oh, A.~H., Agarwal, A., Belgrave, D., and Cho, K. (eds.),
  \emph{Advances in Neural Information Processing Systems}, 2022.
\newblock URL \url{https://openreview.net/forum?id=TJUNtiZiTKE}.

\bibitem[Xu et~al.(2022)Xu, Yu, Song, Shi, Ermon, and Tang]{xu2022geodiff}
Xu, M., Yu, L., Song, Y., Shi, C., Ermon, S., and Tang, J.
\newblock Geodiff: A geometric diffusion model for molecular conformation
  generation.
\newblock \emph{arXiv preprint arXiv:2203.02923}, 2022.

\bibitem[Yu et~al.(2022)Yu, Li, Koh, Zhang, Pang, Qin, Ku, Xu, Baldridge, and
  Wu]{yu2022vectorquantized}
Yu, J., Li, X., Koh, J.~Y., Zhang, H., Pang, R., Qin, J., Ku, A., Xu, Y.,
  Baldridge, J., and Wu, Y.
\newblock Vector-quantized image modeling with improved {VQGAN}.
\newblock In \emph{International Conference on Learning Representations}, 2022.
\newblock URL \url{https://openreview.net/forum?id=pfNyExj7z2}.

\bibitem[Zang \& Wang(2020)Zang and Wang]{zang2020moflow}
Zang, C. and Wang, F.
\newblock Moflow: an invertible flow model for generating molecular graphs.
\newblock In \emph{Proceedings of the 26th ACM SIGKDD International Conference
  on Knowledge Discovery \& Data Mining}, pp.\  617--626, 2020.

\bibitem[Zeng et~al.(2022)Zeng, Vahdat, Williams, Gojcic, Litany, Fidler, and
  Kreis]{zeng2022lion}
Zeng, X., Vahdat, A., Williams, F., Gojcic, Z., Litany, O., Fidler, S., and
  Kreis, K.
\newblock {LION}: Latent point diffusion models for 3d shape generation.
\newblock In Oh, A.~H., Agarwal, A., Belgrave, D., and Cho, K. (eds.),
  \emph{Advances in Neural Information Processing Systems}, 2022.
\newblock URL \url{https://openreview.net/forum?id=tHK5ntjp-5K}.

\end{thebibliography}
